\documentclass[11pt]{article}
\usepackage{inconsolata}
\usepackage[final]{acl}

\usepackage{times}
\usepackage{latexsym}
\usepackage{amssymb}

\usepackage{fontspec}
\usepackage{xeCJK}
\usepackage{polyglossia}

\setdefaultlanguage{english}
\setotherlanguages{hebrew,arabic,russian,ukrainian,bulgarian}

\newfontfamily\hebrewfont{FreeSerif.ttf}[
    Script=Hebrew,
    Extension      = .ttf ,
    BoldFont       = FreeSerifBold ,
    ItalicFont     = FreeSerifItalic ,
    BoldItalicFont = FreeSerifBoldItalic
]
\newfontfamily\hebrewfonttt{FreeSerif.ttf}[
    Script=Hebrew,
    Extension      = .ttf ,
    BoldFont       = FreeSerifBold ,
    ItalicFont     = FreeSerifItalic ,
    BoldItalicFont = FreeSerifBoldItalic
]
\newfontfamily\arabicfont{FreeSerif.ttf}[
    Script=Arabic,
    Extension      = .ttf ,
    BoldFont       = FreeSerifBold ,
    ItalicFont     = FreeSerifItalic ,
    BoldItalicFont = FreeSerifBoldItalic
]
\newfontfamily\arabicfonttt{FreeSerif.ttf}[
    Script=Arabic,
    Extension      = .ttf ,
    BoldFont       = FreeSerifBold ,
    ItalicFont     = FreeSerifItalic ,
    BoldItalicFont = FreeSerifBoldItalic
]

\newfontfamily\russianfont{FreeSerif.ttf}[
    Script=Cyrillic,
    Extension      = .ttf ,
    BoldFont       = FreeSerifBold ,
    ItalicFont     = FreeSerifItalic ,
    BoldItalicFont = FreeSerifBoldItalic
]
\newfontfamily\ukrainianfont{FreeSerif.ttf}[
    Script=Cyrillic,
    Extension      = .ttf ,
    BoldFont       = FreeSerifBold ,
    ItalicFont     = FreeSerifItalic ,
    BoldItalicFont = FreeSerifBoldItalic
]
\newfontfamily\bulgarianfont{FreeSerif.ttf}[
    Script=Cyrillic,
    Extension      = .ttf ,
    BoldFont       = FreeSerifBold ,
    ItalicFont     = FreeSerifItalic ,
    BoldItalicFont = FreeSerifBoldItalic
]

\setCJKsansfont{FandolHei-Regular.otf}

\newCJKfontfamily\zhfont{FandolSong-Regular.otf}
\newCJKfontfamily\jafont{HaranoAjiMincho-Regular.otf}
\newCJKfontfamily\kofont{UnBatang.ttf}

\newfontfamily\devanagarifont{FreeSerif.ttf}[
    Script=Devanagari,
    Extension      = .ttf ,
    BoldFont       = FreeSerifBold ,
    ItalicFont     = FreeSerifItalic ,
    BoldItalicFont = FreeSerifBoldItalic
]
\newfontfamily\bengalifont{FreeSerif.ttf}[
    Script=Bengali,
    Extension      = .ttf ,
    BoldFont       = FreeSerifBold ,
    ItalicFont     = FreeSerifItalic ,
    BoldItalicFont = FreeSerifBoldItalic
]
\newfontfamily\thaifont{FreeSerif.ttf}[
    Script=Thai,
    Extension      = .ttf ,
    BoldFont       = FreeSerifBold ,
    ItalicFont     = FreeSerifItalic ,
    BoldItalicFont = FreeSerifBoldItalic
]

\newcommand{\langHE}[1]{\texthebrew{#1}}
\newcommand{\langAR}[1]{\textarabic{#1}}
\newcommand{\langRU}[1]{\textrussian{#1}}
\newcommand{\langUK}[1]{\textukrainian{#1}}
\newcommand{\langBG}[1]{\textbulgarian{#1}}
\newcommand{\langZH}[1]{{\zhfont #1}}
\newcommand{\langJA}[1]{{\jafont #1}}
\newcommand{\langKO}[1]{{\kofont #1}}

\newenvironment{CJK}[3]{}{}
\newenvironment{CJK*}[3]{}{}
\usepackage{microtype}
\usepackage{dsfont}

\usepackage{inconsolata}

\usepackage{graphicx}
\usepackage{booktabs}
\usepackage{multirow}
\usepackage{amsmath}
\usepackage{cleveref}

\usepackage{tikz}
\usepackage{xcolor}
\usetikzlibrary{arrows.meta, positioning}
\usepackage{placeins}

\title{Cross-Lingual Exploration for Parametric Knowledge}

\author{
  Elisha Diskind\textsuperscript{1},
  Itamar Trainin\textsuperscript{1},
  Uri Shaham\textsuperscript{2},
  Leshem Choshen\textsuperscript{3},
  Idan Szpektor\textsuperscript{2},
  Omri Abend\textsuperscript{1} \\
  \textsuperscript{1}The Hebrew University of Jerusalem \\
  \textsuperscript{2}Google Research \\
  \textsuperscript{3}MIT, IBM-MIT Watson AI Lab \\
  \texttt{\{mosheel.diskind,itamar.trainin,omri.abend\}@mail.huji.ac.il} \\
  \texttt{\{urishaham,szpektor\}@google.com}
}

\begin{document}
\maketitle

\begin{abstract}


Parametric knowledge in Large Language Models is not equally accessible across languages. 
As a result, standard inference techniques often struggle to surface localized facts, leading to failures in cross-lingual knowledge transfer and consistency. 
In this work, we investigate techniques for accessing hidden factual knowledge by exploring cross-lingual prompting strategies.  
We identify four inherent dimensions of cross-lingual exploration that directly govern parametric knowledge retrieval and evaluate them on multilingual factual benchmarks covering 17 typologically diverse languages.
Our results demonstrate that cross-lingual exploration significantly improves knowledge transfer and factual recall, representing a more efficient compute Pareto frontier than native-language scaling. 
Furthermore, we observe corresponding improvements in cross-lingual consistency, exceeding what can be explained by accuracy gains alone. 
Overall, our work establishes multilingual prompt exploration as a highly effective inference-time strategy for unlocking latent parametric knowledge. 

\end{abstract}

\section{Introduction}

A defining characteristic of Large Language Models (LLMs) lies in their ability to encode massive amounts of real-world, multi-lingual and multi-cultural knowledge in their parameters \citep{petroni2019language,brown2020language,ventura2025navigating}.
This parametric knowledge is not only foundational for natural interaction but it further provides a scalable repository of information that can complement or exceed human specialty \citep{guo2024embracing, Singhal2022LargeLM}. Yet, the mechanisms by which this knowledge is stored, accessed, and manipulated remain poorly understood. It is intricately encoded in parametric superposition \citep{templeton2024scaling} across different components of the network \citep{geva2021transformer,meng2022locating,geva2023dissecting}.
Consequently, when presented with a factual question, correct answers often remain unreachable for standard inference methods  \citep{gekhman2025inside}. We show that suitable inference, leveraging multi-lingual exploration, can mitigate this gap.

\begin{figure}[!t]
  \centering
  \includegraphics[width=0.48\textwidth]{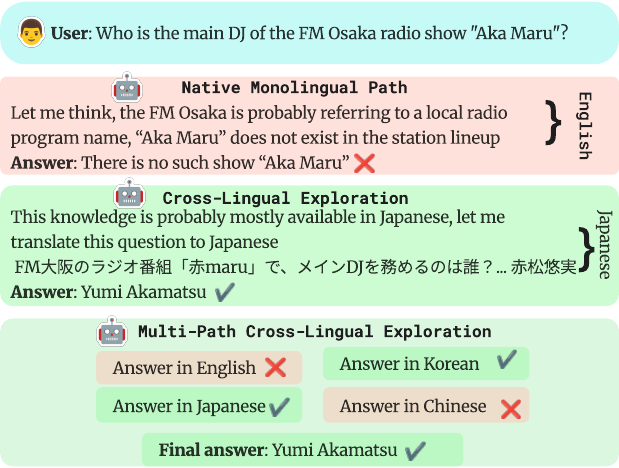}
  \caption{Illustration of cross-lingual exploration for parametric knowledge retrieval. On the top, a monolingual English reasoning path hallucinates an incorrect answer, whereas in the middle, reasoning through Japanese accesses the relevant localized knowledge and retrieves the correct entity. At the bottom, aggregating multiple linguistic paths further increases the probability of retrieving the correct answer.}
  \label{fig:fig1}
\end{figure}

Multilingual settings are a case in point, as factual recall varies substantially across languages and scripts \citep{jiang2020xfactr,kassner2021multilingual,fierro2022factual}. The same underlying fact may thus be easier to access in one language, script, or cultural context than another.

These access failures motivate the evaluation of cross-lingual knowledge transfer and cross-lingual consistency. The former asks whether knowledge acquired or most readily accessible in one language can answer equivalent queries in another, the latter asks whether semantically equivalent questions across languages produce the same factual answer \citep{fierro2022factual,qi2023cross,ifergan2025beneath}. Consistency is often a desideratum of independent value from accuracy, as it provides robustness downstream applications may build on.

Existing inference-time approaches aim to improve multilingual performance by changing the input or reasoning language, typically through translation to a fixed high-resource pivot such as English or through cross-lingual prompting \citep{qin2023crosslingual,etxaniz2024think,mondshine2025beyond}. Recent work on Language Specific Knowledge further challenges the fixed-English assumption, showing that some queries are better answered in non-English expert languages and proposing language selection as a way to improve question answering \citep{agarwal2025language}. However, our goal is broader than identifying an expert language for culturally grounded question answering. Indeed, we characterize the principal dimensions of cross-lingual exploration and analyze their impact on cross-lingual knowledge transfer and factual recall. We further demonstrate that these improvements cannot be attributed solely to the additional computation introduced by exploration. Finally, we show that cross-lingual exploration yields stronger cross-lingual consistency beyond what is explained through accuracy gains alone.

\begin{figure}[!t]
  \centering
  \includegraphics[width=0.49\textwidth]{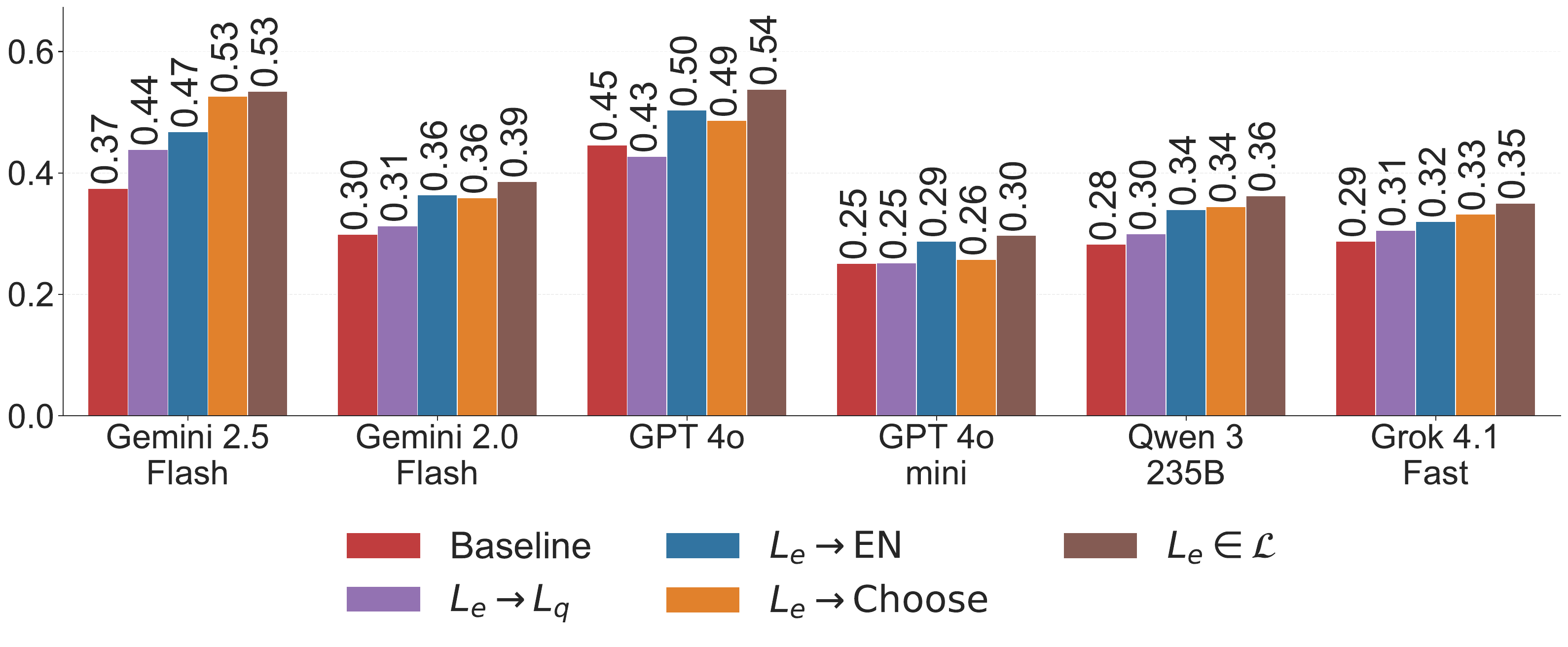}
  \caption{Knowledge Transfer reported for selected strategy implementations. Cross-lingual exploration consistently improves performance across all evaluated models. See \cref{tab:exploration_strategies} for a list of all exploration strategies.}
  \label{fig:eclektic_multi_model_kt_comparison}
\end{figure}

\section{Related Work}

\subsection{Exploration for Parametric Knowledge Retrieval}

Previous work used reasoning techniques \citep{wei2022chainofthought} to improve the performance of language models though the exploration of the solution space. Similarly, \citet{wang2023selfconsistency} improves self-consistency by sampling and aggregating multiple reasoning paths. Recent work connects this idea directly to factual recall. \citet{gekhman2025inside} shows that models can contain hidden factual knowledge that is not expressed in their outputs, and \citet{gekhman2026thinking} shows that reasoning can help recover factual answers that direct prompting fails to produce. These findings suggest that factual recall depends not only on what is stored in the model, but also on whether an inference technique can access it. Our work builds upon this, positioning the language as the central axis of the inference-time exploration.



\subsection{Multilinguality at Inference-Time}




Multilingual prompting and translation-based inference have been used to improve model performance across languages. \citet{qin2023crosslingual} introduces cross-lingual prompting and cross-lingual self-consistency, showing that multilingual reasoning paths can improve zero-shot chain-of-thought reasoning. \citet{etxaniz2024think} study self-translation into English and show that routing non-English inputs through English can often improve performance. \citet{mondshine2025beyond}  showed that the benefit of translation depends on which prompt components are translated and on properties of the source and target languages. 

Recent work further challenges the assumption that English is always the best inference language.
\citet{agarwal2025language} show that, especially for culture-specific knowledge, reasoning in a non-English language can outperform English.
Rather than treating multilingual inference as a task-specific prompting method, we frame cross-lingual exploration as a multi-dimensional design space that unlocks hidden parametric knowledge and expands the evaluation scope to budget-aware cross-lingual knowledge transfer and consistency, isolating the precise benefits of cross-lingual exploration.

\begin{table*}[t]
    \centering
    \small
    \begin{tabular}{lp{4cm}lll}
    \toprule
    \textbf{Strategy} & \textbf{Language Selection} & \textbf{Structure} & \textbf{Aggregation} & \textbf{Budget} \\
    \midrule
    $L_e \rightarrow \mathrm{EN}$ & Fixed Pivot (EN) & Single-path & None & Low \\
    $L_e \rightarrow \mathrm{Choose\text{-}Unlimited}$ & Autonomous & Single-path & None & Low \\
    $L_e \rightarrow\mathrm{Choose\text{-}ReasonAboutOrigin}$ & Autonomous (Origin-aware) & Single-path & None & Low \\
    $L_e \rightarrow \mathrm{Choose} + \mathrm{EN}$ & Autonomous + Fixed Pivot & Sequential & None & Med \\
    $L_e \in \mathcal{L}$ & Fixed Set ($\mathcal{L}$) & Multi-path & Majority / Minority & High \\
    $L_e \in \mathcal{L}_{\mathrm{pred}}$ & Predicted Subset & Multi-path & Majority / Minority & Med \\
    $L_e \in \mathrm{Choose}$ (x13) & Repeated Autonomous & Multi-path & Majority / Minority & High \\
    \midrule
    \multicolumn{5}{c}{\textit{Baselines \& Ablations}} \\
    \midrule
    Native Baseline & Original ($L_q$) & Direct Answer & None & Low \\
    $L_e \rightarrow L_q$ & Original ($L_q$) & Single-path & None & Low \\
    Native (x13) & Original ($L_q$) & Multi-path & Majority / Minority & High \\
    $L_e \rightarrow L_{\mathrm{source}}$ (Oracle) & Oracle Source & Single-path & None & Low \\
    \bottomrule
    \end{tabular}
    \caption{Summary of cross-lingual exploration strategies across four design dimensions. \textbf{Single-path} strategies use a fixed English pivot ($L_e \rightarrow \mathrm{EN}$) or autonomous model-driven selection ($\mathrm{Choose}$), where \textit{Origin-aware} selection requires prior reasoning about the fact's cultural source. \textbf{Hybrid} exploration sequentially traverses in more than one language in a single-path. \textbf{Multi-path} strategies aggregate independent reasoning trajectories using Majority-Vote or Minority-Aware selection across a fixed set ($\mathcal{L}$), a predicted subset ($\mathcal{L}_{\mathrm{pred}}$), or repeated autonomous choices (x13). \textbf{Baselines} include direct retrieval, monolingual reasoning ($L_e \rightarrow L_q$), and budget-matched native ensembles, alongside an oracle source-language routing.}
    \label{tab:exploration_strategies}
\end{table*}

\section{Cross-Lingual Exploration for Knowledge Recall}

To better understand the contributing factors that improve cross-lingual exploration, we decompose the methodological design space into four inherent dimensions (\S\ref{sec:dim_of_exp}). These dimensions govern how, and at what computational cost, knowledge is retrieved during inference, framing cross-lingual exploration as a structured search process for unlocking hidden parametric knowledge. Building on this framework, we then use these dimensions to controllably implement a suite of effective exploration strategies (\S\ref{sec:exp_strategies}).

\subsection{Dimensions of Multilingual Parametric Knowledge Exploration} \label{sec:dim_of_exp}

\paragraph{Language Selection.}
This dimension determines the linguistic contexts used to probe the model's parameters. Since factual knowledge is unevenly distributed across languages, the choice of exploration language is critical for successful retrieval. Different strategies can use internal model properties, such as to route queries toward culture-specific regions of the parametric space or to leverage the observation that models often better represent knowledge in high-resource languages. 

\paragraph{Exploration Routing.}

This dimension defines the routing of the search process. By encouraging the model to traverse different linguistic paths, the exploration structure determines whether the search is narrow and focused on a single linguistic context or broad and distributed across multiple independent viewpoints. Variations include a single-path trajectory that commits to one language throughout the exploration, multiple independent paths that allow different linguistic contexts to compete or complement each other, and sequential language-switching within a single path.

\paragraph{Answer Selection and Aggregation.}
When exploration produces multiple candidate trajectories, this dimension governs how the final factual output is distilled. It represents the mechanism for resolving potential conflicts between linguistic contexts and selecting the most reliable answer. This step is a critical bottleneck for realized performance, while broad exploration increases the probability that the correct fact is surfaced in at least one path, the aggregation strategy must successfully identify it among competing (and potentially incorrect) alternatives. Strategies can range from simple majority voting to more complex, reasoning-aware selection methods.

\paragraph{Inference Budget.}
This dimension accounts for the total compute allocated to the search, the number of model calls, generated tokens, and parallel paths required by a strategy. Budget tracking is critical, as computational cost remains a primary bottleneck in LLM inference. Moreover, controlling for this dimension allows to isolate the intrinsic benefits of uncovering parametric knowledge from the raw performance gains driven purely by general inference-time scaling.

\subsection{Exploration Strategies} \label{sec:exp_strategies}


Building upon this design space, we construct a suite of strategies that probe the model's parametric space through different linguistic trajectories. Formally, we define an exploration strategy $S$ as a function $S(q, L_q) \to a$ that maps a factual question $q$ in input language $L_q$ to a final answer $a$. Each strategy is composed of: (i) a selection mechanism that identifies a set of exploration languages $\mathcal{L}_e \subseteq \mathcal{L}$; (ii) a routing process that generates an exploration trajectory $T_{L_e}$ for each $L_e \in \mathcal{L}_e$, and (iii) an aggregation function $A$ that distills these trajectories into the final answer $a = A(\{T_{L_e} \mid L_e \in \mathcal{L}_e\})$. By varying these components, we instantiate strategies that represent specific configurations within the exploration design space. These strategies are summarized in \cref{tab:exploration_strategies}.

\section{Experimental Setup}

\subsection{Evaluation Tasks and Datasets}

To robustly evaluate the impact of cross-lingual exploration on accessing parametric knowledge, we use two complementary benchmarks.

\paragraph{ECLeKTic:} 
The ECLeKTic (Evaluating Cross-Lingual Knowledge Transfer) dataset \citep{goldman2025eclektic} measures the effectiveness of an LLM in transferring localized knowledge across languages. It targets "language-specific" facts, information corresponding to entities that possess a \textit{Wikipedia} article in one source language but lack equivalents in the other evaluated languages. These facts are instantiated as natural language questions. To succeed on this benchmark, a model must be able to retrieve a fact learned primarily in its source language and accurately answer a query posed in a target language where the fact was never explicitly encountered.

\paragraph{CLIKE:}
While the previous focuses on localized facts, CLIKE (Cross-Lingual Knowledge Editing) dataset \citep{ifergan2025beneath} provides a broad evaluation of general \textit{Wikidata} facts across 13 typologically diverse languages. Each fact in CLIKE is structured as a subject-relation-object triplet and instantiated into natural language cloze-style queries. Because the same underlying fact is queried across all languages, CLIKE allows us to simultaneously measure overall factual recall and cross-lingual answer consistency.


Due to computational costs, we conduct our analysis on a representative subset of $2K$ CLIKE entities, yielding approximately $18K$ multilingual queries. This subset is sufficiently large to ensure a robust evaluation of factual recall and consistency across all available languages, covering roughly $234K$ exploration paths for multi-path strategies.

Our evaluation spans over 13 languages for CLIKE and 12 for ECLeKTic (a total of 17 unique languages), covering diverse scripts and language families. See Appendix \ref{sec:language_details} for details.

\subsection{Models}
We evaluate our strategies across a diverse set of state-of-the-art LLMs, including Gemini 2.5 Flash, Gemini 2.0 Flash \citep{gemini25report2025}, GPT-4o \citep{openai2024gpt4o}, GPT-4o-mini \citep{openai2024gpt4omini}, Qwen 3 235B \citep{qwen2025qwen3}, and Grok 4.1 Fast \citep{xai2025grok41fast}. This suite of models covers both proprietary and open-source models, as well as different model sizes. Knowledge cutoffs for all models are reported in Appendix \ref{sec:model_details}. Knowledge cutoff dates are particularly relevant for the ECLeKTic benchmark to prevent contamination of the gold answers in the model parameters. Detailed hyperparameters and configurations are provided in Appendix \ref{sec:exp_details}.

\subsection{Strategy Instantiation}

To instantiate these strategies, we develop a suite of prompt templates tailored for each strategy and language, as detailed in Appendix~\ref{sec:prompt_templates}. A core design principle of our experimental setup is that the full prompt, including instructions for translation, intermediate reasoning, and answer formatting is always provided in the original query language $L_q$. For example, if a question is posed in Japanese and the strategy requires exploring a trajectory in English, the instruction ``Translate the following question to English, answer it, and then translate the final answer back to Japanese'' is itself presented in Japanese. This approach avoids introducing uncontrolled code-switching within the prompt and isolates the model's ability to explore its own internal parametric space using its native cross-lingual capabilities.

\subsection{Evaluation Metrics}

\paragraph{Accuracy} We evaluate the correctness of model predictions using Gemini 2.5 Flash \citep{gemini25report2025} as an LLM-as-a-judge, tasked with classifying whether the prediction is equivalent to the gold answer (using the prompt provided in Appendix \ref{sec:evaluation_prompts}), regardless of phrasing. We report mean accuracy with Standard Error of the Mean (SEM), and both Macro (language-level) and Micro (sample-level) averages.

\paragraph{Knowledge-Transfer}
We quantify cross-lingual knowledge transfer ($KT$) by measuring the model's ability to correctly answer queries from the ECLeKTic dataset in a target language, where the underlying fact is localized to a different source language. Formally, let $\mathcal{D}_{EC}$ be the ECLeKTic dataset where each sample is a quadruplet $(q, L_q, L_{src}, a_{gold})$, with $q$ being the question, $L_q$ the query language, $L_{src}$ the language where the fact is uniquely localized, and $a_{gold}$ the gold answer. We define Knowledge-Transfer for a strategy $S$ as:
\begin{equation}
KT(S) = \frac{1}{|\mathcal{D}|} \sum_{\mathcal{D}} \mathds{1}[S(q, L_q) = a_{gold}]
\end{equation}
where, $\mathcal{D}\!=\!\{(q,\!L_q,\!L_{src},\!a_{gold})\!\in\!\mathcal{D}_{EC}\!\mid\!L_q\!\neq\! L_{src}\}$

\paragraph{Factual Recall}
We define Factual Recall ($FR$) as the model's ability to correctly retrieve and express factual knowledge that is nominally present within the query language corpus, unlike $KT$, which involves retrieving facts localized to a distinct source language. This metric allows us to distinguish between bridging cross-lingual knowledge gaps and recalling facts that are likewise present in the query language. In our experiments, we use accuracy on the CLIKE benchmark as the primary measure of $FR$.

\paragraph{Upper Bounds}
For all multi-path methods, we compute a potential upper bounds (e.g., Potential - $L_e \in \mathcal{L}$). In this setting, an example is considered successful if at least one generated path contains the correct answer. This decouples the effectiveness of exploration from the bottlenecks of the aggregation strategy, providing an estimate of the knowledge unlocked by the search. However, we note that this upper bound may be optimistic, as taking the maximum over multiple stochastic generation paths can inflate performance estimates even when gains are due to noise.

\paragraph{Inference Budget Measurement}
To compare the computational cost of different strategies on an even ground, we measure the token cost of each method. The cost is computed as the average total tokens consumed per question, encompassing both the input tokens and the generated response output tokens. For multi-path strategies, the cost is the sum of tokens across all generated paths and the aggregation step.

\begin{figure*}[!t]
  \centering
  \includegraphics[width=1\textwidth]{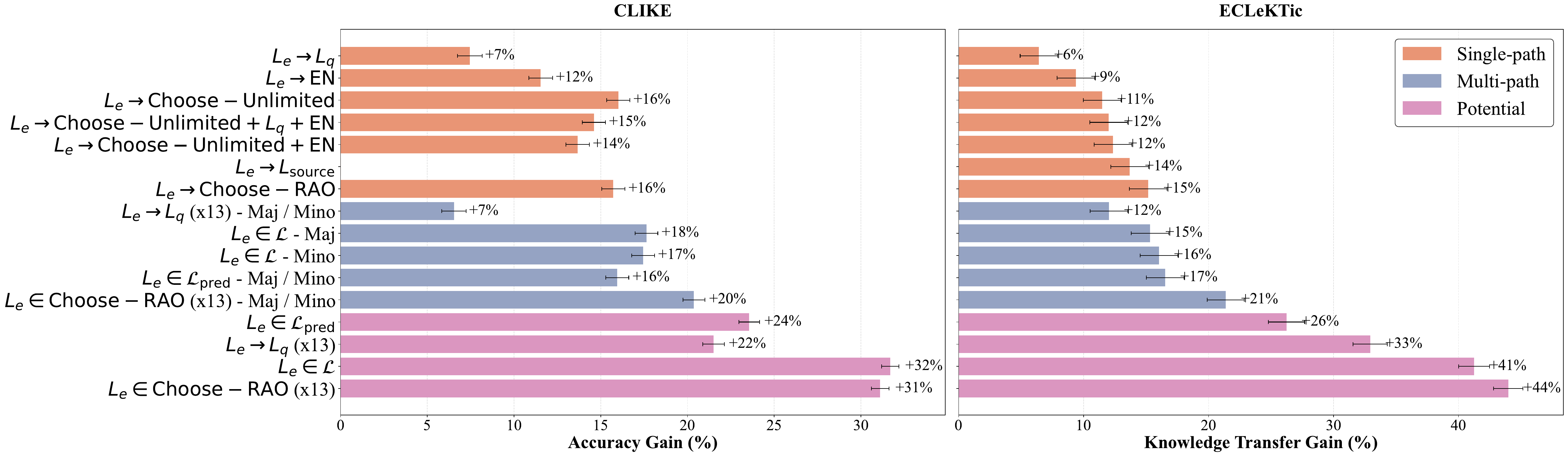}
  \caption{Accuracy gain ($\Delta_{FR}={FR}_{\text{native}}-{FR}_{S}$) on CLIKE dataset (left) and Knowledge Transfer gain ($\Delta_{KT}={KT}_{\text{native}}-{KT}_{S}$) on ECLeKTic (right) relative to the Native Baseline, across exploration strategies. Strategy names are abbreviated: Maj (Majority Vote), Mino (Minority Aware), and RAO (ReasonAboutOrigin). For methods with two aggregation labels (e.g., Maj / Mino), the first label applies to CLIKE and the second to ECLeKTic.}
  \label{fig:clike_accuracy_gain_and_eclektic_knowledge_transfer_gain}
\end{figure*}

\section{Results}

\subsection{Exploration Improves Knowledge Transfer}

Our techniques utilize alternative linguistic contexts to encourage wider exploration of the model's parametric knowledge and improve the recall of localized facts. To evaluate this, we measure knowledge transfer ($KT$) on the ECLeKTic dataset. As shown in \cref{fig:eclektic_multi_model_kt_comparison}, our exploration strategies consistently improve $KT$ across all evaluated models. Furthermore, \cref{fig:clike_accuracy_gain_and_eclektic_knowledge_transfer_gain} indicates that every cross-lingual variant achieves a positive $KT$ gain of up to $21\%$ compared to the native monolingual baseline. Crucially, these gains occur without any updates to the model parameters or access to external information. These results demonstrate that cross-lingual exploration can successfully surface localized facts that were previously inaccessible through standard querying methods.

\subsection{Exploration Improves Factual Recall}
We investigate whether cross-lingual exploration additionally improves general factual recall. Specifically, we evaluate facts that are nominally present within the query language corpus. To evaluate this, we use the CLIKE benchmark, where every queried fact is represented in \textit{Wikipedia} across all evaluated languages. As shown in \cref{fig:clike_accuracy_gain_and_eclektic_knowledge_transfer_gain}, cross-lingual exploration significantly boosts factual recall across all evaluated methods. This directly indicates that parametric facts that are present in the query language but are inaccessible by standard monolingual strategies benefit from cross-lingual exploration. This goes beyond bridging knowledge transfer as a fact may be represented in the query language itself. Overall, this shows that traversing alternative linguistic paths provides a robust mechanism for closing the "recall gap" even in low-budget, single-path routing.

\subsection{Separating Reasoning from Cross-Lingual Exploration}

Our results demonstrate the clear advantage of cross-lingual exploration, we now verify that these gains genuinely stem from uncovering hidden parametric knowledge through exploration rather than merely from inference-time scaling via reasoning. To isolate this effect, we compared our cross-lingual exploration strategies against a monolingual baseline that utilizes an intermediate thinking path (reasoning) in the original query language ($L_e \rightarrow L_q$). As shown in \cref{fig:clike_accuracy_gain_and_eclektic_knowledge_transfer_gain}, although monolingual reasoning yields a modest performance boost ($+7\%$ on CLIKE and $+6\%$ on ECLeKTic), it is substantially outperformed by cross-lingual routing ($L_e \rightarrow \mathrm{Choose\text{-}ReasonAboutOrigin}$), which at the same cost achieves gains of $+16\%$ and $+15\%$, respectively. This substantial margin demonstrates that the language shift rather than the reasoning process itself, is the primary mechanism for unlocking previously inaccessible parametric knowledge.

\section{Impact of Dimensions}

\subsection{The Selection Dimension}

The choice of exploration language ($L_e$) dictates the specific region of the parametric space being probed. A frequent assumption in the literature is that routing through a high-resource pivot, such as English, yields optimal retrieval. Our results challenge this assumption.

While English-only exploration ($L_e \rightarrow \mathrm{EN}$) provides a substantial improvement over the native baseline (e.g., $12\%$ on CLIKE and $9\%$ on ECLeKTic), it falls short of autonomous selection strategies, specifically, $L_e \rightarrow \mathrm{Choose\text{-}ReasonAboutOrigin}$ achieves improvements of $16\%$ and $15\%$ respectively, outperforming the English pivot (\cref{fig:clike_accuracy_gain_and_eclektic_knowledge_transfer_gain}). This gap suggests that a single pivot language cannot act as a universal key to all latent knowledge, as facts localized in specific cultural contexts often require traversal through their associated linguistic representations to be successfully retrieved.

The autonomous \textbf{origin-aware selection} ($53.1\%$) \textbf{surpasses even the provided oracle} setting ($L_e \rightarrow L_{\mathrm{source}}$, $51.0\%$), where the model is explicitly routed through the fact's original source language provided in the ECLeKTic dataset. 
This result reinforces recent evidence that multilingual models possess some ability to select useful reasoning languages on their own through their LLM-selected language baseline \citep{agarwal2025language}. We show that autonomous selection is not only competitive with fixed or externally supplied language choices, but can potentially outperform routing based on provided source-language metadata.
This suggests that having the model first reason about the source language it should explore in, greatly improves its ability to surface relevant knowledge, and potentially more effective than supplying the true source language in advance.

To study the underlying mechanism of autonomous selection, we performed a qualitative analysis of reasoning traces where English-pivot reasoning failed but autonomous language selection succeeded on the ECLeKTic benchmark. As summarized in Appendix~\ref{sec:trace_analysis_examples}, we observe that while English-pivot paths often default to generic hallucinations or incorrect heuristics for localized queries, autonomous selection allows the model to identify the entity's cultural origin and retrieve precise factual details by switching to the associated language.


\subsection{The Routing Dimension} 

The specific routing of cross-lingual exploration greatly impacts its effectiveness. Our results in \cref{fig:clike_accuracy_gain_and_eclektic_knowledge_transfer_gain} show a clear performance hierarchy, while a single-path autonomous selection strategy ($L_e \rightarrow \mathrm{Choose\text{-}ReasonAboutOrigin}$) already yields substantial gains (e.g., $+15.7\%$ on CLIKE), extending to multi-path exploration ($L_e \in \mathrm{Choose\text{-}ReasonAboutOrigin}$ (x13)) consistently further improves accuracy by $4.7\%$ and $6.2\%$ on CLIKE and ECLeKTic, respectively. This demonstrates that multi-path exploration are superior for parametric retrieval, and aligns with existing work reporting that exploration across multiple independent paths improves parametric retrieval \citep{wang2023selfconsistency}.

In contrast, we find that a hybrid routing (i.e., $L_e \rightarrow \mathrm{Choose} + \mathrm{EN}$), where the model transitions through multiple languages within a single generation path does not contribute to performance gains over $L_e \rightarrow \mathrm{Choose\text{-}Unlimited}$. To investigate this, we used an LLM-as-a-judge (Gemini 2.5 Pro) to evaluate whether the model's semantic answer changes across the different languages within the same path (see extraction prompt in Appendix~\ref{sec:evaluation_prompts}). Our analysis shows that in $99\%$ of cases, the answer remained identical after the initial language shift. This indicates that sequential switching provides no additional retrieval benefit while consuming a larger token expense. 
Consequently, \textbf{parallel multi-path exploration or a single optimized autonomous path represent the most effective routing strategies}.
 
\subsection{The Aggregation Dimension}

Our analysis reveals a massive gap between the potential upper bound and the knowledge successfully selected during aggregation. For instance, On the CLIKE benchmark with the $L_e \in \mathcal{L}$ strategy, we report potential knowledge recall gain of over $31\%$  relative to the native baseline, while the realized gain is approximately $18\%$ (\cref{fig:clike_accuracy_gain_and_eclektic_knowledge_transfer_gain}). Similarly on the ECLeKTic benchmark, the potential knowledge transfer gain exceeds $44\%$, compared to a realized gain of approximately $21\%$. This shows that cross-lingual exploration successfully reaches and unlocks latent parametric knowledge in the vast majority of cases, while identifying the correct answer among competing candidates remains a significant challenge.

The nature of the knowledge may dictate the optimal aggregation strategy. In CLIKE, where facts are broadly represented across languages, the delta achieved by Majority-Vote is slightly higher than that of Minority-Aware aggregation ($+17.6\%$ vs. $+17.4\%$ ). However, on the ECLeKTic benchmark, which targets highly localized knowledge, Minority-Aware aggregation takes a lead over Majority Vote ($+16.0\% $ vs. $+15.3\%$). This may occur because correct parametric associations for localized facts might only be triggered in a narrow set of languages, potentially making the correct answer a minority among hallucinated responses from other paths.

\subsection{Cross-Lingual Exploration as an Efficient Compute Frontier}

\begin{figure}[!t]
    \centering
    \includegraphics[width=\linewidth]{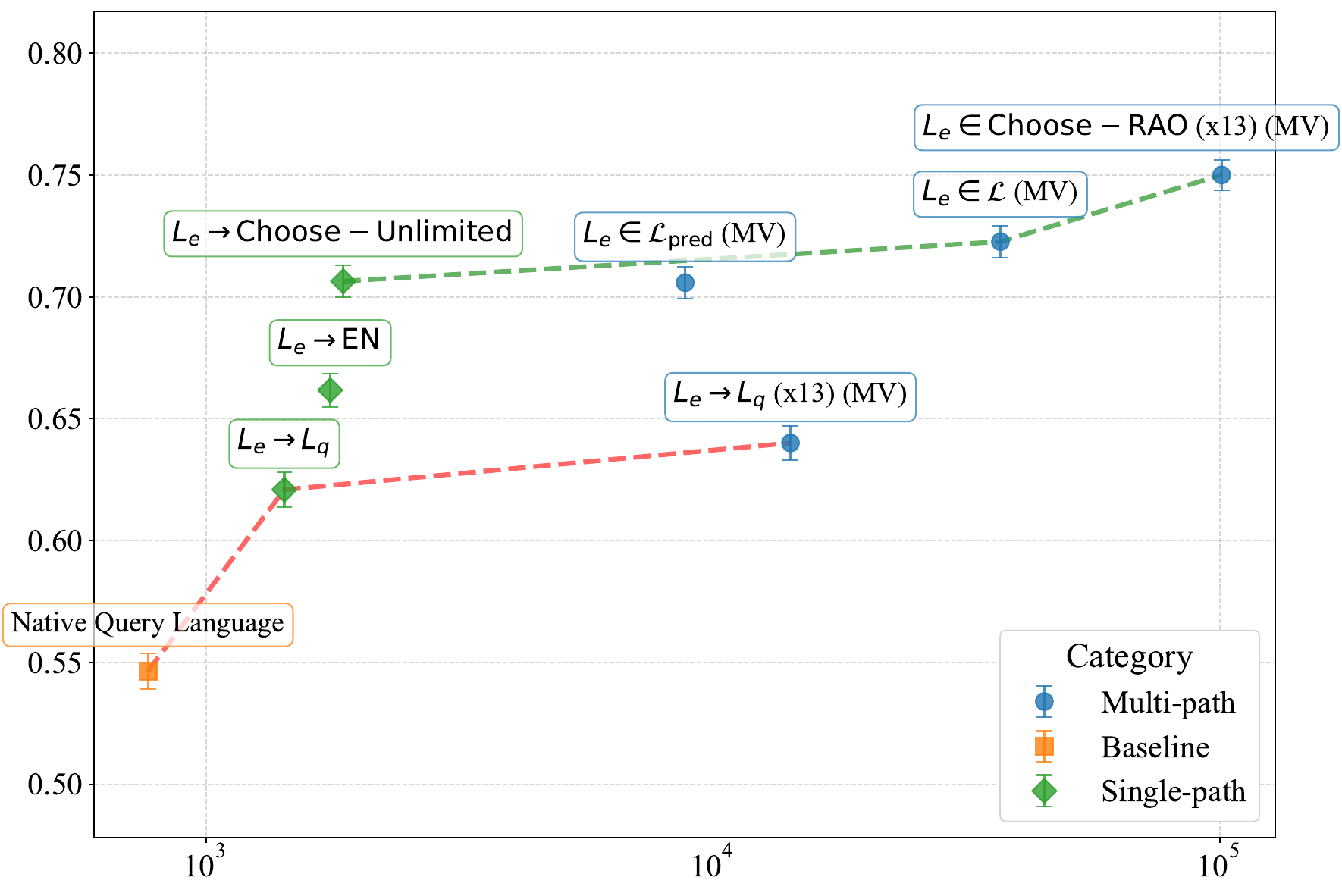}
    \caption{Accuracy as a function of Average Token Cost (log scale) on the CLIKE benchmark. The green dashed curve (cross-lingual exploration) dominates the red dashed curve (query language scaling), demonstrating that traversing alternative linguistic paths is a more efficient use of inference-time compute than scaling within the native query language.}
    \label{fig:clike_budget}
\end{figure}

A central question in inference-time scaling is whether performance gains stem from additional computation or from a more effective search. We evaluate this by comparing the Pareto frontier of accuracy versus cost for native-language scaling and cross-lingual exploration. As shown in \Cref{fig:clike_budget}, the red Pareto curve, which corresponds to scaling compute within the native query language, presents low efficiency in factual recall. In contrast, the green Pareto curve, capturing cross-lingual exploration, outperforms the native language at every point. The Pareto frontier for ECLeKTic, which shows a comparable pattern, is provided in Appendix \ref{sec:appendix_eclektic_budget}.

\section{Measuring Cross-Lingual Consistency}
\label{sec:consistency-method}


Cross-lingual consistency is a key desideratum for robust LLMs, ensuring that factual queries yield the same answer regardless of input language. This stability is valuable independently of accuracy, as it provides predictable behavior for downstream applications. However, consistency is often confounded with accuracy, as a perfectly accurate model is necessarily perfectly consistent. To isolate intrinsic consistency improvements from accuracy gains, we formalize a hypothesis testing framework to compare any two methods. We test the null hypothesis ($H_0$) that consistency gains are strictly a side-effect of improved accuracy, against the alternative ($H_1$) that a method intrinsically produces more consistent answers, even when controlling for whether those answers are correct or incorrect.

\subsection{Variables and Definitions}
For any pair of answers to a given query across a language pair ($L_1$, $L_2$), we define
$C$ (Consistency) to be a binary outcome, where $C = 1$ if the LLM judges the two answers as semantically consistent, and $0$ otherwise.
$M$ (Method) is a binary indicator comparing two methods, where $M = 0$ for a baseline method and $M = 1$ for an evaluated experimental method.
Finally, $S$ (Joint Accuracy State) is a categorical variable representing the correctness of the pair, taking one of three values: \emph{Both Correct} ($A_1=1 \land A_2=1$), \emph{One Correct} ($A_1 \neq A_2$), or \emph{Both Incorrect} ($A_1=0 \land A_2=0$).

\subsection{Hypothesis Testing}
Our null hypothesis states that consistency ($C$) is conditionally independent of the method ($M$) given the joint accuracy state ($S$). Formally, $P(C=1 | S, M=0) = P(C=1 | S, M=1)$. Under $H_0$, a new method only increases overall consistency by shifting the distribution of $S$ toward the {\it Both Correct} state.
We employ three primary statistical methods to test this:

\paragraph{Macro Analysis (Expected vs. Observed):} We compute the expected consistency of the experimental method ($M=1$) under $H_0$ by applying the conditional consistency rates from the baseline ($M=0$) to the experimental method's joint accuracy distribution: 
\begin{equation}
\sum_{s \in S} P_{M=0}(C=1 | S=s) \cdot P_{M=1}(S=s) 
\end{equation}
We then compare this expected rate against the actual observed consistency of $M=1$.

\paragraph{Cochran-Mantel-Haenszel (CMH) Test:} We stratify the data by the three accuracy states $S$ to test for the conditional independence of $C$ and $M$ \citep{mantel1959statistical}. A rejection of the null hypothesis indicates an intrinsic consistency effect beyond accuracy gains.

\paragraph{Logistic Regression:} To quantify the effect size, we fit a logistic regression model \citep{hosmer2013applied}: $\text{logit}(P(C=1)) = \beta_0 + \beta_1 I_{S=\text{One Correct}} + \beta_2 I_{S=\text{Both Correct}} + \beta_3 M$. The coefficient $\beta_3$ represents the log odds ratio of consistency for the experimental method versus the baseline, holding accuracy constant.

\subsection{Results}
All cross-lingual strategies yield significant intrinsic consistency gains ($p < 0.001$ for both CMH and Logistic Regression). For instance, the Multi-Path strategy  ($L_e \in \mathcal{L}$) achieves a $6$ point gain beyond its expected consistency relative to the native baseline. Similarly, Single-path ($L_e \rightarrow \mathrm{Choose\text{-}Unlimited}$) shows a significant intrinsic gain of $4.3\%$. Furthermore, Multi-Path exploration outperforms the Single-Path reasoning baseline ($L_e \rightarrow L_q$) by $4.5\%$ in intrinsic consistency. We conclude that cross-lingual exploration improves consistency beyond what can be explained by accuracy gains alone. Detailed results are provided in Table~\ref{tab:hypothesis_results} in Appendix~\ref{sec:appendix_consistency_results}.

\section{Conclusion}

While LLMs store immense knowledge in their parameters, some of it may be inaccessible to standard inference methods. We show the effectiveness of cross-lingual exploration to overcome this challenge. By isolating the underlying mechanism that specific facts may be accessible in one language while remaining latent in others, we demonstrate that transitioning between languages during inference, improves access to knowledge that is underexplored using conventional parametric retrieval techniques. We define a four dimensional design space of inference methods, to systematically compare different options. Our analysis shows substantial improvement in factual recall, knowledge transfer, and intrinsic consistency beyond simple accuracy gains, while maintaining a more efficient compute-performance trade-off than native-language scaling. 

We observe that the answer aggregation dimension is a primary bottleneck, marking a key area for future research. Overall, our results suggest that explicitly training models to strategically perform cross-lingual exploration could be a promising direction for building more robust and factually accessible multilingual LLMs. By characterizing the dimensions of cross-lingual exploration, our work provides a structured path toward more reliable multilingual systems.

\section*{Limitations}
This work has several limitations. First, our evaluation relies on automatic LLM-based judgments for correctness and consistency. This allows evaluation at the scale required by multi-path cross-lingual exploration, but may introduce errors, especially for answers involving aliases or partial entity names. To mitigate this concern, we manually verified 100 randomly sampled examples across various languages and observed that the judge model classifies the answers correctly

Second, while our empirical analysis covers two factual knowledge benchmarks suited for studying cross-lingual factual access, we do not cover all forms of multilingual such as cross-lingual complex problem-solving.

Finally, our experiments depend partly on proprietary or externally hosted models whose internal architectures and training protocols are undisclosed. While we utilize specific model versions to improve reproducibility, the "black-box" nature of these systems and the potential for service-side deprecations or undocumented updates remain inherent risks to long-term stability.

\bibliography{custom}

\onecolumn
\clearpage
\appendix

\section{Cross-Lingual Exploration Across Methods and Languages}

\begin{center}
\captionsetup{type=table}
\centering
\small
\resizebox{\textwidth}{!}{
\begin{tabular}{l|ccccccccccccc||cc}
\toprule
\textsc{Method} & \multicolumn{4}{c|}{\textbf{Latin}} & \multicolumn{3}{c|}{\textbf{Cyrillic}} & \multicolumn{2}{c|}{\textbf{Devanagari}} & \multicolumn{1}{c|}{\textbf{Chinese}} & \multicolumn{1}{c|}{\textbf{Japanese}} & \multicolumn{1}{c|}{\textbf{Hebrew}} & \multicolumn{1}{c|}{\textbf{Arabic}} & \textbf{Macro} & \textbf{Micro} \\
 & \texttt{EN} & \texttt{FR} & \texttt{IT} & \texttt{ES} & \texttt{RU} & \texttt{UK} & \texttt{BG} & \texttt{HI} & \texttt{BN} & \texttt{ZH} & \texttt{JA} & \texttt{HE} & \texttt{AR} & \textsc{Lang Avg} & \textsc{Global} \\
\midrule
\multicolumn{16}{l}{\textbf{Baselines}} \\
\midrule
\hspace{0.3cm} $L_e \rightarrow L_q$ (x13) - Naive & 72.1 \tiny{$\pm$1.0} & 71.9 \tiny{$\pm$1.0} & 69.1 \tiny{$\pm$1.1} & 72.2 \tiny{$\pm$1.0} & 59.8 \tiny{$\pm$1.1} & 66.3 \tiny{$\pm$1.4} & 75.2 \tiny{$\pm$2.0} & 71.4 \tiny{$\pm$3.3} & 54.4 \tiny{$\pm$2.4} & 50.4 \tiny{$\pm$1.2} & 56.5 \tiny{$\pm$1.2} & 61.2 \tiny{$\pm$1.4} & 53.0 \tiny{$\pm$1.3} & 64.1 \tiny{$\pm$0.4} & 64.0 \tiny{$\pm$0.4} \\
\hspace{0.3cm} $L_e \rightarrow L_q$ & 72.6 \tiny{$\pm$1.0} & 70.7 \tiny{$\pm$1.0} & 67.3 \tiny{$\pm$1.1} & 66.2 \tiny{$\pm$1.1} & 58.1 \tiny{$\pm$1.1} & 60.5 \tiny{$\pm$1.4} & 73.5 \tiny{$\pm$2.1} & 63.5 \tiny{$\pm$3.5} & 44.8 \tiny{$\pm$2.4} & 53.4 \tiny{$\pm$1.2} & 52.6 \tiny{$\pm$1.2} & 62.2 \tiny{$\pm$1.4} & 52.1 \tiny{$\pm$1.3} & 61.3 \tiny{$\pm$0.4} & 62.1 \tiny{$\pm$0.4} \\
\hspace{0.3cm} $L_e \rightarrow L_q$ (x13) - Majority Vote & 70.0 \tiny{$\pm$1.0} & 69.5 \tiny{$\pm$1.0} & 67.3 \tiny{$\pm$1.1} & 70.0 \tiny{$\pm$1.0} & 55.7 \tiny{$\pm$1.1} & 62.2 \tiny{$\pm$1.4} & 74.1 \tiny{$\pm$2.1} & 67.2 \tiny{$\pm$3.4} & 52.5 \tiny{$\pm$2.4} & 48.6 \tiny{$\pm$1.2} & 53.3 \tiny{$\pm$1.2} & 54.4 \tiny{$\pm$1.4} & 50.9 \tiny{$\pm$1.3} & 61.2 \tiny{$\pm$0.4} & 61.2 \tiny{$\pm$0.4} \\
\hspace{0.3cm} Native Query Language & 66.5 \tiny{$\pm$1.1} & 64.4 \tiny{$\pm$1.1} & 63.8 \tiny{$\pm$1.1} & 66.5 \tiny{$\pm$1.1} & 50.2 \tiny{$\pm$1.1} & 53.6 \tiny{$\pm$1.5} & 68.2 \tiny{$\pm$2.2} & 65.1 \tiny{$\pm$3.5} & 44.4 \tiny{$\pm$2.4} & 37.3 \tiny{$\pm$1.2} & 39.3 \tiny{$\pm$1.2} & 49.4 \tiny{$\pm$1.4} & 43.3 \tiny{$\pm$1.3} & 54.8 \tiny{$\pm$0.4} & 54.6 \tiny{$\pm$0.4} \\
\midrule
\multicolumn{16}{l}{\textbf{Single-path}} \\
\midrule
\hspace{0.3cm} $L_e \rightarrow \mathrm{Choose-Unlimited}$ & 74.3 \tiny{$\pm$1.0} & 74.1 \tiny{$\pm$1.0} & 72.3 \tiny{$\pm$1.0} & 75.0 \tiny{$\pm$1.0} & 69.9 \tiny{$\pm$1.1} & 73.2 \tiny{$\pm$1.3} & 80.0 \tiny{$\pm$1.9} & 67.2 \tiny{$\pm$3.4} & 64.7 \tiny{$\pm$2.3} & 58.7 \tiny{$\pm$1.2} & 70.6 \tiny{$\pm$1.1} & 70.3 \tiny{$\pm$1.3} & 64.7 \tiny{$\pm$1.3} & 70.4 \tiny{$\pm$0.4} & 70.6 \tiny{$\pm$0.3} \\
\hspace{0.3cm} $L_e \rightarrow \mathrm{Choose-ReasonAboutOrigin}$ & \underline{75.5} \tiny{$\pm$1.0} & 74.9 \tiny{$\pm$1.0} & 72.4 \tiny{$\pm$1.0} & \underline{75.2} \tiny{$\pm$1.0} & 67.5 \tiny{$\pm$1.1} & 74.5 \tiny{$\pm$1.3} & 79.8 \tiny{$\pm$1.9} & 64.0 \tiny{$\pm$3.5} & 64.5 \tiny{$\pm$2.3} & 57.1 \tiny{$\pm$1.2} & 66.8 \tiny{$\pm$1.1} & 70.2 \tiny{$\pm$1.3} & \underline{67.0} \tiny{$\pm$1.2} & 70.0 \tiny{$\pm$0.4} & 70.4 \tiny{$\pm$0.3} \\
\hspace{0.3cm} $L_e \rightarrow \mathrm{Choose-Unlimited} + L_q + \mathrm{EN}$ & 73.4 \tiny{$\pm$1.0} & 73.6 \tiny{$\pm$1.0} & 71.9 \tiny{$\pm$1.0} & 74.1 \tiny{$\pm$1.0} & 68.9 \tiny{$\pm$1.1} & 73.8 \tiny{$\pm$1.3} & 79.4 \tiny{$\pm$1.9} & 64.6 \tiny{$\pm$3.5} & 58.0 \tiny{$\pm$2.4} & 54.9 \tiny{$\pm$1.2} & 66.8 \tiny{$\pm$1.1} & 68.1 \tiny{$\pm$1.3} & 65.3 \tiny{$\pm$1.3} & 68.7 \tiny{$\pm$0.4} & 69.2 \tiny{$\pm$0.3} \\
\hspace{0.3cm} $L_e \rightarrow \mathrm{Choose-Unlimited} + \mathrm{EN}$ & 75.2 \tiny{$\pm$1.0} & 74.6 \tiny{$\pm$1.0} & 70.6 \tiny{$\pm$1.0} & 72.6 \tiny{$\pm$1.0} & 65.7 \tiny{$\pm$1.1} & 71.4 \tiny{$\pm$1.3} & 77.9 \tiny{$\pm$1.9} & 65.1 \tiny{$\pm$3.5} & 61.2 \tiny{$\pm$2.4} & 56.6 \tiny{$\pm$1.2} & 65.9 \tiny{$\pm$1.1} & 65.7 \tiny{$\pm$1.4} & 60.2 \tiny{$\pm$1.3} & 67.9 \tiny{$\pm$0.4} & 68.3 \tiny{$\pm$0.3} \\
\hspace{0.3cm} $L_e \rightarrow \mathrm{EN}$ & 66.5 \tiny{$\pm$1.1} & 70.5 \tiny{$\pm$1.0} & 68.1 \tiny{$\pm$1.1} & 70.0 \tiny{$\pm$1.0} & 63.8 \tiny{$\pm$1.1} & 71.1 \tiny{$\pm$1.3} & 79.2 \tiny{$\pm$1.9} & 70.4 \tiny{$\pm$3.3} & 61.2 \tiny{$\pm$2.4} & 54.3 \tiny{$\pm$1.2} & 67.2 \tiny{$\pm$1.1} & 67.9 \tiny{$\pm$1.3} & 58.8 \tiny{$\pm$1.3} & 66.8 \tiny{$\pm$0.4} & 66.2 \tiny{$\pm$0.4} \\
\midrule
\multicolumn{16}{l}{\textbf{Multi-path}} \\
\midrule
\hspace{0.3cm} $L_e \in \mathrm{Choose-ReasonAboutOrigin}$ (x13) - Majority Vote & \textbf{77.7} \tiny{$\pm$0.9} & \textbf{78.3} \tiny{$\pm$0.9} & \textbf{76.5} \tiny{$\pm$1.0} & \textbf{77.8} \tiny{$\pm$0.9} & \textbf{73.6} \tiny{$\pm$1.0} & \textbf{77.2} \tiny{$\pm$1.2} & \textbf{83.1} \tiny{$\pm$1.8} & 65.6 \tiny{$\pm$3.5} & \textbf{69.1} \tiny{$\pm$2.3} & \textbf{66.6} \tiny{$\pm$1.2} & \textbf{73.8} \tiny{$\pm$1.1} & \textbf{76.8} \tiny{$\pm$1.2} & \textbf{70.9} \tiny{$\pm$1.2} & \textbf{74.4} \tiny{$\pm$0.4} & \textbf{75.0} \tiny{$\pm$0.3} \\
\hspace{0.3cm} $L_e \in \mathcal{L}$ - Majority Vote & 74.1 \tiny{$\pm$1.0} & \underline{75.5} \tiny{$\pm$1.0} & \underline{75.1} \tiny{$\pm$1.0} & \underline{75.2} \tiny{$\pm$1.0} & \underline{70.4} \tiny{$\pm$1.0} & \underline{75.0} \tiny{$\pm$1.3} & 81.4 \tiny{$\pm$1.8} & \underline{74.1} \tiny{$\pm$3.2} & 65.0 \tiny{$\pm$2.3} & \underline{63.4} \tiny{$\pm$1.2} & \underline{72.1} \tiny{$\pm$1.1} & \underline{73.3} \tiny{$\pm$1.3} & 66.2 \tiny{$\pm$1.2} & \underline{72.4} \tiny{$\pm$0.4} & \underline{72.3} \tiny{$\pm$0.3} \\
\hspace{0.3cm} $L_e \in \mathcal{L}$ - Naive & 73.1 \tiny{$\pm$1.0} & 74.7 \tiny{$\pm$1.0} & 73.8 \tiny{$\pm$1.0} & 73.6 \tiny{$\pm$1.0} & 68.9 \tiny{$\pm$1.1} & 74.0 \tiny{$\pm$1.3} & \underline{82.0} \tiny{$\pm$1.8} & \textbf{74.6} \tiny{$\pm$3.2} & 61.9 \tiny{$\pm$2.4} & 62.9 \tiny{$\pm$1.2} & 70.1 \tiny{$\pm$1.1} & 71.8 \tiny{$\pm$1.3} & 64.6 \tiny{$\pm$1.3} & 71.2 \tiny{$\pm$0.4} & 71.0 \tiny{$\pm$0.3} \\
\hspace{0.3cm} $L_e \in \mathcal{L}_{\mathrm{pred}}$ - Majority Vote & 73.9 \tiny{$\pm$1.0} & 73.8 \tiny{$\pm$1.0} & 73.2 \tiny{$\pm$1.0} & 74.1 \tiny{$\pm$1.0} & 69.6 \tiny{$\pm$1.1} & 74.0 \tiny{$\pm$1.3} & 80.9 \tiny{$\pm$1.8} & 65.1 \tiny{$\pm$3.5} & \underline{65.5} \tiny{$\pm$2.3} & 60.0 \tiny{$\pm$1.2} & 69.1 \tiny{$\pm$1.1} & 70.7 \tiny{$\pm$1.3} & 64.6 \tiny{$\pm$1.3} & 70.4 \tiny{$\pm$0.4} & 70.6 \tiny{$\pm$0.3} \\
\hspace{0.3cm} $L_e \in \mathcal{L}$ - Extracted Answer & 73.3 \tiny{$\pm$1.0} & 73.2 \tiny{$\pm$1.0} & 73.0 \tiny{$\pm$1.0} & 74.1 \tiny{$\pm$1.0} & 66.9 \tiny{$\pm$1.1} & 70.3 \tiny{$\pm$1.3} & 79.4 \tiny{$\pm$1.9} & 73.0 \tiny{$\pm$3.2} & 59.2 \tiny{$\pm$2.4} & 58.1 \tiny{$\pm$1.2} & 67.4 \tiny{$\pm$1.1} & 68.3 \tiny{$\pm$1.3} & 59.0 \tiny{$\pm$1.3} & 68.9 \tiny{$\pm$0.4} & 68.8 \tiny{$\pm$0.3} \\
\midrule
\multicolumn{16}{l}{\textbf{Upper Bounds}} \\
\midrule
\hspace{0.3cm} $L_e \in \mathcal{L}$ & 87.2 \tiny{$\pm$0.7} & 87.4 \tiny{$\pm$0.7} & 87.9 \tiny{$\pm$0.7} & 87.9 \tiny{$\pm$0.7} & 85.1 \tiny{$\pm$0.8} & 87.8 \tiny{$\pm$1.0} & 91.7 \tiny{$\pm$1.3} & 93.7 \tiny{$\pm$1.8} & 82.0 \tiny{$\pm$1.9} & 80.1 \tiny{$\pm$1.0} & 86.2 \tiny{$\pm$0.8} & 88.5 \tiny{$\pm$0.9} & 83.7 \tiny{$\pm$1.0} & 86.9 \tiny{$\pm$0.3} & 86.3 \tiny{$\pm$0.3} \\
\hspace{0.3cm} $L_e \in \mathrm{Choose-ReasonAboutOrigin}$ (x13) & 86.8 \tiny{$\pm$0.8} & 86.9 \tiny{$\pm$0.8} & 86.9 \tiny{$\pm$0.8} & 87.5 \tiny{$\pm$0.7} & 85.4 \tiny{$\pm$0.8} & 87.6 \tiny{$\pm$1.0} & 91.7 \tiny{$\pm$1.3} & 78.3 \tiny{$\pm$3.0} & 82.7 \tiny{$\pm$1.9} & 79.7 \tiny{$\pm$1.0} & 85.3 \tiny{$\pm$0.8} & 86.7 \tiny{$\pm$1.0} & 84.3 \tiny{$\pm$1.0} & 85.4 \tiny{$\pm$0.3} & 85.7 \tiny{$\pm$0.3} \\
\hspace{0.3cm} $L_e \in \mathcal{L}_{\mathrm{pred}}$ & 81.1 \tiny{$\pm$0.9} & 80.7 \tiny{$\pm$0.9} & 81.1 \tiny{$\pm$0.9} & 81.6 \tiny{$\pm$0.9} & 76.9 \tiny{$\pm$1.0} & 81.0 \tiny{$\pm$1.2} & 87.5 \tiny{$\pm$1.6} & 77.2 \tiny{$\pm$3.1} & 74.1 \tiny{$\pm$2.1} & 68.0 \tiny{$\pm$1.1} & 77.5 \tiny{$\pm$1.0} & 77.7 \tiny{$\pm$1.2} & 72.9 \tiny{$\pm$1.2} & 78.3 \tiny{$\pm$0.4} & 78.2 \tiny{$\pm$0.3} \\
\hspace{0.3cm} $L_e \rightarrow L_q$ (x13) & 84.0 \tiny{$\pm$0.8} & 83.2 \tiny{$\pm$0.8} & 81.5 \tiny{$\pm$0.9} & 84.2 \tiny{$\pm$0.8} & 72.3 \tiny{$\pm$1.0} & 78.3 \tiny{$\pm$1.2} & 85.5 \tiny{$\pm$1.6} & 83.6 \tiny{$\pm$2.7} & 69.8 \tiny{$\pm$2.3} & 63.6 \tiny{$\pm$1.2} & 68.5 \tiny{$\pm$1.1} & 70.6 \tiny{$\pm$1.3} & 66.9 \tiny{$\pm$1.2} & 76.3 \tiny{$\pm$0.4} & 76.1 \tiny{$\pm$0.3} \\
\bottomrule
\end{tabular}
}
\caption{Factual Knowledge Recall Accuracy across different languages and reasoning strategies for CLIKE. Results are reported as Mean $\pm$ SEM in \%. Best results (excluding Potential upper bound) are \textbf{bolded} and second-best are \underline{underlined}. Methods are grouped by their reasoning paradigm (Baselines, Single-path, Multi-path).}
\label{tab:res_clike}
\end{center}

\vspace{1em}

\begin{center}
\captionsetup{type=table}
\centering
\small
\resizebox{\textwidth}{!}{
\begin{tabular}{l|cccccccccccc||cc}
\toprule
\textsc{Method} & \multicolumn{7}{c|}{\textbf{Latin}} & \multicolumn{1}{c|}{\textbf{Devanagari}} & \multicolumn{1}{c|}{\textbf{Chinese}} & \multicolumn{1}{c|}{\textbf{Japanese}} & \multicolumn{1}{c|}{\textbf{Korean}} & \multicolumn{1}{c|}{\textbf{Hebrew}} & \textbf{Macro} & \textbf{Micro} \\
 & \texttt{EN} & \texttt{FR} & \texttt{IT} & \texttt{ES} & \texttt{PT} & \texttt{DE} & \texttt{ID} & \texttt{HI} & \texttt{ZH} & \texttt{JA} & \texttt{KO} & \texttt{HE} & \textsc{Lang Avg} & \textsc{Global} \\
\midrule
\multicolumn{15}{l}{\textbf{Baselines}} \\
\midrule
\hspace{0.3cm} $L_e \rightarrow L_q$ (x13) - Minority Aware & 53.9 \tiny{$\pm$2.5} & 55.2 \tiny{$\pm$2.5} & 53.1 \tiny{$\pm$2.5} & 55.7 \tiny{$\pm$2.5} & 53.4 \tiny{$\pm$2.5} & 52.9 \tiny{$\pm$2.6} & 54.4 \tiny{$\pm$2.5} & 40.9 \tiny{$\pm$2.5} & 53.9 \tiny{$\pm$2.5} & 45.6 \tiny{$\pm$2.5} & 45.1 \tiny{$\pm$2.5} & 43.2 \tiny{$\pm$2.5} & 50.6 \tiny{$\pm$0.7} & 50.6 \tiny{$\pm$0.7} \\
\hspace{0.3cm} $L_e \rightarrow L_q$ & 49.2 \tiny{$\pm$2.6} & 50.0 \tiny{$\pm$2.6} & 46.4 \tiny{$\pm$2.5} & 44.0 \tiny{$\pm$2.5} & 43.0 \tiny{$\pm$2.5} & 50.3 \tiny{$\pm$2.6} & 48.7 \tiny{$\pm$2.6} & 38.8 \tiny{$\pm$2.5} & 48.7 \tiny{$\pm$2.6} & 39.3 \tiny{$\pm$2.5} & 34.4 \tiny{$\pm$2.4} & 42.7 \tiny{$\pm$2.5} & 44.6 \tiny{$\pm$0.7} & 44.6 \tiny{$\pm$0.7} \\
\hspace{0.3cm} Native Query Language & 44.5 \tiny{$\pm$2.5} & 41.4 \tiny{$\pm$2.5} & 42.7 \tiny{$\pm$2.5} & 39.6 \tiny{$\pm$2.5} & 41.1 \tiny{$\pm$2.5} & 47.4 \tiny{$\pm$2.6} & 40.6 \tiny{$\pm$2.5} & 29.7 \tiny{$\pm$2.3} & 36.7 \tiny{$\pm$2.5} & 35.4 \tiny{$\pm$2.4} & 29.9 \tiny{$\pm$2.3} & 32.0 \tiny{$\pm$2.4} & 38.4 \tiny{$\pm$0.7} & 38.4 \tiny{$\pm$0.7} \\
\midrule
\multicolumn{15}{l}{\textbf{Single-path}} \\
\midrule
\hspace{0.3cm} $L_e \rightarrow \mathrm{Choose-ReasonAboutOrigin}$ & \underline{56.2} \tiny{$\pm$2.5} & 54.4 \tiny{$\pm$2.5} & 56.0 \tiny{$\pm$2.5} & 57.0 \tiny{$\pm$2.5} & 55.5 \tiny{$\pm$2.5} & 54.2 \tiny{$\pm$2.5} & 54.4 \tiny{$\pm$2.5} & 48.7 \tiny{$\pm$2.6} & 55.5 \tiny{$\pm$2.5} & 50.3 \tiny{$\pm$2.6} & 45.8 \tiny{$\pm$2.5} & 49.7 \tiny{$\pm$2.6} & 53.1 \tiny{$\pm$0.7} & 53.1 \tiny{$\pm$0.7} \\
\hspace{0.3cm} $L_e \rightarrow L_{\mathrm{source}}$ & 52.9 \tiny{$\pm$2.6} & 49.2 \tiny{$\pm$2.6} & 50.8 \tiny{$\pm$2.6} & 54.7 \tiny{$\pm$2.5} & 51.6 \tiny{$\pm$2.6} & 54.4 \tiny{$\pm$2.5} & 52.9 \tiny{$\pm$2.6} & 47.4 \tiny{$\pm$2.6} & 50.0 \tiny{$\pm$2.6} & \textbf{54.9} \tiny{$\pm$2.5} & 44.8 \tiny{$\pm$2.5} & 48.2 \tiny{$\pm$2.6} & 51.0 \tiny{$\pm$0.7} & 51.0 \tiny{$\pm$0.7} \\
\hspace{0.3cm} $L_e \rightarrow \mathrm{Choose-Unlimited} + \mathrm{EN}$ & 53.1 \tiny{$\pm$2.5} & 53.6 \tiny{$\pm$2.5} & 51.3 \tiny{$\pm$2.6} & 52.1 \tiny{$\pm$2.6} & 50.3 \tiny{$\pm$2.6} & 54.4 \tiny{$\pm$2.5} & 55.7 \tiny{$\pm$2.5} & 46.1 \tiny{$\pm$2.5} & 50.5 \tiny{$\pm$2.6} & 44.5 \tiny{$\pm$2.5} & 45.3 \tiny{$\pm$2.5} & 47.4 \tiny{$\pm$2.6} & 50.4 \tiny{$\pm$0.7} & 50.4 \tiny{$\pm$0.7} \\
\hspace{0.3cm} $L_e \rightarrow \mathrm{Choose-Unlimited} + L_q + \mathrm{EN}$ & 53.6 \tiny{$\pm$2.5} & 55.7 \tiny{$\pm$2.5} & 49.5 \tiny{$\pm$2.6} & 54.2 \tiny{$\pm$2.5} & 53.9 \tiny{$\pm$2.5} & 53.1 \tiny{$\pm$2.5} & 54.7 \tiny{$\pm$2.5} & 45.3 \tiny{$\pm$2.5} & 50.0 \tiny{$\pm$2.6} & 45.1 \tiny{$\pm$2.5} & 40.6 \tiny{$\pm$2.5} & 45.6 \tiny{$\pm$2.5} & 50.1 \tiny{$\pm$0.7} & 50.1 \tiny{$\pm$0.7} \\
\hspace{0.3cm} $L_e \rightarrow \mathrm{Choose-Unlimited}$ & 54.2 \tiny{$\pm$2.5} & 54.4 \tiny{$\pm$2.5} & 50.5 \tiny{$\pm$2.6} & 53.9 \tiny{$\pm$2.5} & 48.7 \tiny{$\pm$2.6} & 53.4 \tiny{$\pm$2.5} & 51.0 \tiny{$\pm$2.6} & 43.8 \tiny{$\pm$2.5} & 49.7 \tiny{$\pm$2.6} & 47.1 \tiny{$\pm$2.6} & 43.8 \tiny{$\pm$2.5} & 45.6 \tiny{$\pm$2.5} & 49.7 \tiny{$\pm$0.7} & 49.7 \tiny{$\pm$0.7} \\
\hspace{0.3cm} $L_e \rightarrow \mathrm{EN}$ & 44.5 \tiny{$\pm$2.5} & 48.7 \tiny{$\pm$2.6} & 50.8 \tiny{$\pm$2.6} & 50.0 \tiny{$\pm$2.6} & 48.2 \tiny{$\pm$2.6} & 51.8 \tiny{$\pm$2.6} & 49.0 \tiny{$\pm$2.6} & 46.4 \tiny{$\pm$2.5} & 49.5 \tiny{$\pm$2.6} & 42.7 \tiny{$\pm$2.5} & 42.4 \tiny{$\pm$2.5} & 47.7 \tiny{$\pm$2.6} & 47.6 \tiny{$\pm$0.7} & 47.6 \tiny{$\pm$0.7} \\
\midrule
\multicolumn{15}{l}{\textbf{Multi-path}} \\
\midrule
\hspace{0.3cm} $L_e \in \mathrm{Choose-ReasonAboutOrigin}$ (x13) - Minority Aware & \textbf{57.8} \tiny{$\pm$2.5} & \textbf{62.8} \tiny{$\pm$2.5} & \underline{57.8} \tiny{$\pm$2.5} & \textbf{61.7} \tiny{$\pm$2.5} & \textbf{61.2} \tiny{$\pm$2.5} & \textbf{62.0} \tiny{$\pm$2.5} & \textbf{63.3} \tiny{$\pm$2.5} & \textbf{55.5} \tiny{$\pm$2.5} & \textbf{62.5} \tiny{$\pm$2.5} & \underline{54.7} \tiny{$\pm$2.5} & \textbf{57.3} \tiny{$\pm$2.5} & \textbf{55.2} \tiny{$\pm$2.5} & \textbf{59.3} \tiny{$\pm$0.7} & \textbf{59.3} \tiny{$\pm$0.7} \\
\hspace{0.3cm} $L_e \in \mathcal{L}_{\mathrm{pred}}$ - Minority Aware & 52.6 \tiny{$\pm$2.6} & \underline{57.8} \tiny{$\pm$2.5} & \textbf{59.1} \tiny{$\pm$2.5} & \underline{58.1} \tiny{$\pm$2.5} & \underline{57.8} \tiny{$\pm$2.5} & 58.3 \tiny{$\pm$2.5} & 56.0 \tiny{$\pm$2.5} & \underline{50.0} \tiny{$\pm$2.6} & \underline{59.6} \tiny{$\pm$2.5} & 51.0 \tiny{$\pm$2.6} & \underline{48.7} \tiny{$\pm$2.6} & 48.2 \tiny{$\pm$2.6} & \underline{54.8} \tiny{$\pm$0.7} & \underline{54.8} \tiny{$\pm$0.7} \\
\hspace{0.3cm} $L_e \in \mathcal{L}$ - Minority Aware & 50.8 \tiny{$\pm$2.6} & 56.0 \tiny{$\pm$2.5} & 54.9 \tiny{$\pm$2.5} & 56.8 \tiny{$\pm$2.5} & 57.6 \tiny{$\pm$2.5} & \underline{60.4} \tiny{$\pm$2.5} & 53.6 \tiny{$\pm$2.5} & 48.4 \tiny{$\pm$2.6} & 56.8 \tiny{$\pm$2.5} & 53.6 \tiny{$\pm$2.5} & 48.2 \tiny{$\pm$2.6} & 50.8 \tiny{$\pm$2.6} & 54.0 \tiny{$\pm$0.7} & 54.0 \tiny{$\pm$0.7} \\
\hspace{0.3cm} $L_e \in \mathcal{L}$ - Majority Vote & 49.2 \tiny{$\pm$2.6} & 56.0 \tiny{$\pm$2.5} & 54.2 \tiny{$\pm$2.5} & 56.2 \tiny{$\pm$2.5} & 56.2 \tiny{$\pm$2.5} & 57.8 \tiny{$\pm$2.5} & \underline{57.0} \tiny{$\pm$2.5} & 48.7 \tiny{$\pm$2.6} & 56.0 \tiny{$\pm$2.5} & 48.2 \tiny{$\pm$2.6} & \underline{48.7} \tiny{$\pm$2.6} & \underline{52.6} \tiny{$\pm$2.6} & 53.4 \tiny{$\pm$0.7} & 53.4 \tiny{$\pm$0.7} \\
\midrule
\multicolumn{15}{l}{\textbf{Upper Bounds}} \\
\midrule
\hspace{0.3cm} $L_e \in \mathrm{Choose-ReasonAboutOrigin}$ (x13) & 83.9 \tiny{$\pm$1.9} & 82.6 \tiny{$\pm$1.9} & 82.8 \tiny{$\pm$1.9} & 84.9 \tiny{$\pm$1.8} & 82.6 \tiny{$\pm$1.9} & 82.8 \tiny{$\pm$1.9} & 83.9 \tiny{$\pm$1.9} & 77.6 \tiny{$\pm$2.1} & 84.1 \tiny{$\pm$1.9} & 78.4 \tiny{$\pm$2.1} & 78.1 \tiny{$\pm$2.1} & 78.9 \tiny{$\pm$2.1} & 81.7 \tiny{$\pm$0.6} & 81.7 \tiny{$\pm$0.6} \\
\hspace{0.3cm} $L_e \in \mathcal{L}$ & 78.6 \tiny{$\pm$2.1} & 80.5 \tiny{$\pm$2.0} & 80.2 \tiny{$\pm$2.0} & 80.5 \tiny{$\pm$2.0} & 80.2 \tiny{$\pm$2.0} & 82.3 \tiny{$\pm$2.0} & 79.7 \tiny{$\pm$2.1} & 75.3 \tiny{$\pm$2.2} & 80.2 \tiny{$\pm$2.0} & 80.2 \tiny{$\pm$2.0} & 76.0 \tiny{$\pm$2.2} & 77.3 \tiny{$\pm$2.1} & 79.3 \tiny{$\pm$0.6} & 79.3 \tiny{$\pm$0.6} \\
\hspace{0.3cm} $L_e \rightarrow L_q$ (x13) & 75.0 \tiny{$\pm$2.2} & 76.8 \tiny{$\pm$2.2} & 76.3 \tiny{$\pm$2.2} & 75.5 \tiny{$\pm$2.2} & 76.8 \tiny{$\pm$2.2} & 77.6 \tiny{$\pm$2.1} & 74.7 \tiny{$\pm$2.2} & 62.8 \tiny{$\pm$2.5} & 73.7 \tiny{$\pm$2.2} & 63.8 \tiny{$\pm$2.5} & 62.2 \tiny{$\pm$2.5} & 63.5 \tiny{$\pm$2.5} & 71.6 \tiny{$\pm$0.7} & 71.6 \tiny{$\pm$0.7} \\
\hspace{0.3cm} $L_e \in \mathcal{L}_{\mathrm{pred}}$ & 64.3 \tiny{$\pm$2.4} & 66.7 \tiny{$\pm$2.4} & 68.5 \tiny{$\pm$2.4} & 66.9 \tiny{$\pm$2.4} & 65.6 \tiny{$\pm$2.4} & 67.4 \tiny{$\pm$2.4} & 66.4 \tiny{$\pm$2.4} & 59.9 \tiny{$\pm$2.5} & 67.2 \tiny{$\pm$2.4} & 62.8 \tiny{$\pm$2.5} & 57.8 \tiny{$\pm$2.5} & 60.4 \tiny{$\pm$2.5} & 64.5 \tiny{$\pm$0.7} & 64.5 \tiny{$\pm$0.7} \\
\bottomrule
\end{tabular}
}
\caption{Factual Knowledge Recall Accuracy across different languages and reasoning strategies for ECLeKTic. Results are reported as Mean $\pm$ SEM in \%. Best results (excluding Potential upper bound) are \textbf{boldfaced} and second-best are \underline{underlined}. Methods are grouped by their reasoning paradigm (Baselines, Single-path, Multi-path).}
\label{tab:res_eclektic}
\end{center}

\newpage

\section{Cross-Lingual Consistency Results} 

\label{sec:appendix_consistency_results}

\begin{table*}[!t]
    \centering
\resizebox{\linewidth}{!}{%
\begin{tabular}{llccccccc}
\toprule
Baseline & Target & Base Cons. & Target Cons. & Exp. Cons. & Gain & CMH $p$ & Logit $\beta$ & Logit $p$ \\
\midrule
Native Query Language & Multi-path, $L\_e \in \mathcal{L}$ & 49.8\% & 74.1\% & 68.1\% & +6.0\% & $<$0.001 & 0.89 & $<$0.001 \\
Native Query Language & Single-path, $L\_e \rightarrow \mathrm{Choose}$ & 49.8\% & 70.8\% & 66.4\% & +4.3\% & $<$0.001 & 0.66 & $<$0.001 \\
Native Query Language & Single-path, $L\_e \rightarrow L\_q$ & 49.8\% & 58.5\% & 56.8\% & +1.7\% & $<$0.001 & 0.23 & $<$0.001 \\
Single-path, $L\_e \rightarrow L\_q$ & Multi-path, $L\_e \in \mathcal{L}$ & 58.5\% & 74.1\% & 69.6\% & +4.5\% & $<$0.001 & 0.65 & $<$0.001 \\
Single-path, $L\_e \rightarrow L\_q$ & Single-path, $L\_e \rightarrow \mathrm{Choose}$ & 58.5\% & 70.8\% & 67.8\% & +2.9\% & $<$0.001 & 0.43 & $<$0.001 \\
\bottomrule
\end{tabular}
}
\caption{Hypothesis Testing Results for Cross-Lingual Consistency. \textbf{Base Cons.} and \textbf{Target Cons.} show the observed cross-lingual consistency for the baseline and target methods, respectively. \textbf{Exp. Cons.} is the expected consistency of the target method under the null hypothesis (i.e., if its consistency gains were solely due to its joint accuracy distribution matching the baseline's conditional consistency rates). \textbf{Gain} denotes the difference between Target Cons. and Exp. Cons. \textbf{CMH $p$} is the p-value from the Cochran-Mantel-Haenszel test stratifying by joint accuracy. \textbf{Logit $\beta$} and \textbf{Logit $p$} represent the coefficient and p-value for the method indicator in a logistic regression modeling consistency, controlling for joint accuracy state.}
\label{tab:hypothesis_results}
\end{table*}

\clearpage

\section{Experimental Details}
\label{sec:exp_details}

\subsection{Language Details}
\label{sec:language_details}

Our evaluation covers 17 unique languages across the CLIKE and ECLeKTic benchmarks. Table \ref{tab:language_details} summarizes the languages, their ISO codes, language families, and scripts.

\begin{table*}[ht]
\centering
\small
\begin{tabular}{lllllc}
\toprule
\textbf{Language} & \textbf{ISO} & \textbf{Family} & \textbf{Script} & \textbf{Region} & \textbf{Datasets} \\
\midrule
English & EN & Germanic (Indo-European) & Latin & Global & CLIKE, ECLeKTic \\
French & FR & Romance (Indo-European) & Latin & Europe/Africa & CLIKE, ECLeKTic \\
Italian & IT & Romance (Indo-European) & Latin & Europe & CLIKE, ECLeKTic \\
Spanish & ES & Romance (Indo-European) & Latin & Europe/Americas & CLIKE, ECLeKTic \\
Portuguese & PT & Romance (Indo-European) & Latin & Europe/Americas & ECLeKTic \\
German & DE & Germanic (Indo-European) & Latin & Europe & ECLeKTic \\
Indonesian & ID & Austronesian & Latin & SE Asia & ECLeKTic \\
Russian & RU & Slavic (Indo-European) & Cyrillic & Europe/Asia & CLIKE \\
Ukrainian & UK & Slavic (Indo-European) & Cyrillic & Europe & CLIKE \\
Bulgarian & BG & Slavic (Indo-European) & Cyrillic & Europe & CLIKE \\
Hindi & HI & Indo-Aryan (Indo-European) & Devanagari & South Asia & CLIKE, ECLeKTic \\
Bengali & BN & Indo-Aryan (Indo-European) & Devanagari & South Asia & CLIKE \\
Chinese & ZH & Sinitic (Sino-Tibetan) & Chinese (Simp.) & East Asia & CLIKE, ECLeKTic \\
Japanese & JA & Japonic & Japanese & East Asia & CLIKE, ECLeKTic \\
Korean & KO & Koreanic & Korean (Hangul) & East Asia & ECLeKTic \\
Hebrew & HE & Semitic (Afroasiatic) & Hebrew & Middle East & CLIKE, ECLeKTic \\
Arabic & AR & Semitic (Afroasiatic) & Arabic & Middle East/Africa & CLIKE \\
\bottomrule
\end{tabular}
\caption{Detailed overview of languages evaluated in this work, including their linguistic classification and coverage in each benchmark.}
\label{tab:language_details}
\end{table*}

\subsection{Hyperparameters and Decoding}
For all generations, we used the default decoding parameters for each model to reflect their standard performance characteristics. For proprietary models (Gemini, GPT-4o, Grok), this typically corresponds to a temperature of 1.0. For the open-weights Qwen 3 model, we used the Google MaaS (Model-as-a-Service) default configuration.

The LLM-as-a-judge were configured with a temperature of 0.0 to maximize reproducibility and ensure deterministic assessments. These models were also instructed to provide outputs in a structured JSON format to facilitate automated parsing.

\subsection{Reasoning and Token Budgets}
Models with native inference-time reasoning capabilities (Gemini 2.5 Flash) were allocated a "thinking budget" of 0 tokens to ensure model output is available for analysis. 

The inference budget reported in Section 5.6 is defined as the average total tokens (input tokens + output tokens) consumed per factual question. For multi-path strategies, the budget represents the aggregate token count across all parallel reasoning paths plus the tokens used for the final aggregation step (e.g., the majority vote or minority-aware selection).

\subsection{Compute Resources}
\label{sec:model_details}
The models evaluated in this work were trained on large-scale industrial compute clusters. For instance, the Gemini 2.5 family was trained on Google's TPUv5p architecture, while the Qwen 3 235B model utilized a diverse dataset of 36 trillion tokens. Our own experiments were conducted using the Gemini Batch API and the OpenAI Batch API.

\subsection{Model Specifications, Sources, and API Access}
To ensure transparency and reproducibility, we provide the technical specifications, original source documentation, and API endpoints for all models evaluated in this work. Table \ref{tab:model_metadata} details the release dates, knowledge cutoffs, and official source links, while Table \ref{tab:api_access} lists the API providers used for inference.

\begin{table*}
\centering
\small
\resizebox{\textwidth}{!}{
\begin{tabular}{llll}
\toprule
\textbf{Model} & \textbf{Release Date} & \textbf{Knowledge Cutoff} & \textbf{Technical Source / Announcement} \\
\midrule
Gemini 2.5 Flash / Pro & June 2025 & January 2025 & \url{https://arxiv.org/abs/2507.06261} \\
Gemini 2.0 Flash-Lite & February 2025 & Late 2024 & \url{https://deepmind.google/technologies/gemini/} \\
GPT-4o & May 2024 & June 2024 & \url{https://openai.com/index/hello-gpt-4o/} \\
GPT-4o-mini & July 2024 & October 2023 & \url{https://openai.com/index/gpt-4o-mini-advancing-cost-efficient-intelligence/} \\
Qwen 3 235B & April 2025 & Early 2025 & \url{https://qwenlm.github.io/blog/qwen3/} \\
Grok 4.1 Fast & November 2025 & Late 2025 & \url{https://x.ai/blog/grok-4.1-fast} \\
\bottomrule
\end{tabular}
}
\caption{Metadata and technical documentation sources for evaluated models.}
\label{tab:model_metadata}
\end{table*}

\begin{table*}
\centering
\small
\resizebox{\textwidth}{!}{
\begin{tabular}{lll}
\toprule
\textbf{Model Family} & \textbf{API Provider} & \textbf{Documentation Link} \\
\midrule
Gemini & Google Vertex AI & \url{https://cloud.google.com/vertex-ai/docs/generative-ai/model-reference/gemini} \\
GPT & OpenAI API & \url{https://platform.openai.com/docs/models/gpt-4o} \\
Qwen / Grok & Vertex AI Model Garden & \url{https://cloud.google.com/vertex-ai/docs/model-garden/explore-models} \\
\bottomrule
\end{tabular}
}
\caption{API endpoints and provider documentation used for inference.}
\label{tab:api_access}
\end{table*}

\subsection{Evaluation Prompts}
\label{sec:evaluation_prompts}

\paragraph{LLM-as-a-Judge (Correctness):}
\textit{You are an expert evaluator. Given a question, a gold target answer, and a model prediction, determine if the model prediction is correct. The model prediction might be in a different language or phrased differently, but if it conveys the same correct meaning or entity as the gold target, it should be considered correct.} \\
\textit{Question: \{original\_question\} \\
Gold Target: \{gold\_answer\} \\
Model Prediction: \{prediction\}} \\
\textit{Respond strictly with a JSON object containing a single boolean field "is\_correct". Example: \texttt{\{"is\_correct": true\}} or \texttt{\{"is\_correct": false\}}}

\paragraph{LLM Consistency Checker:}
\textit{You are an expert multilingual evaluator. You will be given a question and two answers provided in potentially different languages. Your task is to determine if the two answers convey the same core information and are consistent with each other.} \\
\textit{Original Question (for context): \{question \} \\
Answer 1 (\{lang1 \}): \{answer1 \} \\
Answer 2 (\{lang2 \}): \{answer2 \}} \\
\textit{Evaluate if the two answers are semantically equivalent or consistently convey the same core fact in the context of the question.}

\paragraph{Single-path Consistency Extraction:}
\textit{Analyze the following model output which contains reasoning and answers in multiple languages. The model was asked to answer a factual question in several languages within a single response. \\
Original Question: \{question\} \\
Model Output: \{prediction\_raw\} \\
Your task: \\
1. Identify all languages in which the model provided an intermediate answer (before the final answer). \\
2. Extract the core answer provided for EACH language. \\
3. Determine if these intermediate answers are consistent with each other. \\
4. Also extract the 'Final Answer' provided at the very end. \\
Respond strictly with a JSON object in the following format: \\
\{ "intermediate\_answers": \{ "lang\_code\_1": "answer\_1", ... \}, "final\_answer": "the\_final\_answer", "is\_consistent\_across\_languages": true/false, "explanation": "..." \}}

\newpage

\section{Qualitative Examples of Cross-Lingual Trajectories}
\label{sec:trace_analysis_examples}

This section provides qualitative examples from the ECLeKTic benchmark where routing through a fixed English pivot failed, but allowing the model to autonomously choose the reasoning language succeeded.

We observe that when forced to route through English, the model frequently produces hallucinated or overly generic answers. For instance, when asked in Japanese about ``Prikon district'', English reasoning hallucinates that the district does not exist, whereas the autonomous strategy correctly selects Spanish (identifying it as related to the Spanish company ``PRECONSA''), and surfaces the precise factual detail. Similarly, for geographically localized queries such as identifying the university of Argentine scholar Rubén Saúl Stiglitz or the ingredients of the traditional Iberian drink ``Gloriao'' (Coffee), the English reasoning path defaults to high-probability but incorrect heuristics (e.g., guessing the University of Buenos Aires for an Argentine lawyer or ``Lemon Zest'' for drinks). In contrast, the autonomous strategy explicitly reasons about the entity's origin, switches to the associated language (e.g., Spanish or Portuguese), and successfully retrieves the localized parametric knowledge.

\begin{table*}
    \centering
    \small
    \begin{tabular}{p{0.15\linewidth} p{0.8\linewidth}}
        \toprule
        \textbf{Original Query} & \begin{CJK}{UTF8}{min}プリコンサ地区はどの会社によって建設されましたか？\end{CJK} (Japanese) \\
        \textbf{Gold Answer} & \begin{CJK}{UTF8}{min}プリコンサ\end{CJK} (PRECONSA) \\
        \midrule
        \textbf{English Path (Failed)} & \textit{Raw Output:} Here's the English translation of your question, followed by the answer and the final response in Japanese: **Question in English:** "Which company built the Prikon district?" **Answer:** There is no readily available information about a "Prikon district" being built by a specific company. It's possible that "Prikon" might be a misspelling... **Final Answer:** \begin{CJK}{UTF8}{min}プリコン地区はどの会社によって建設されたかについての情報は見つかりませんでした。\end{CJK} \\
        \midrule
        \textbf{Chosen Path (Success)} & \textit{Raw Output:} The language in which information about the Preconsa district is most likely to be found is Spanish. **Spanish Translation:** '¿Por qué empresa fue construido el barrio de Preconsa?' **Answer in Spanish:** El barrio de Preconsa fue construido por la empresa Promotora de Construcciones, S.A. (PRECONSA). **Final Answer in Japanese:** \begin{CJK}{UTF8}{min}プリコンサ地区は、プロモトラ・デ・コンストルクシオネス S.A.（PRECONSA）社によって建設されました。\end{CJK} \\
        \bottomrule
    \end{tabular}
    \caption{Example showing the model identifying the Spanish origins of a district mentioned in a Japanese query, enabling successful retrieval that failed in English.}
\end{table*}

\begin{table*}
    \centering
    \small
    \begin{tabular}{p{0.15\linewidth} p{0.8\linewidth}}
        \toprule
        \textbf{Original Query} & In quale università Rubén Saúl Stiglitz ha conseguito la laurea in Giurisprudenza? (Italian) \\
        \textbf{Gold Answer} & Università Nazionale di La Plata \\
        \midrule
        \textbf{English Path (Failed)} & \textit{Raw Output:} Ecco la traduzione della domanda, la risposta in inglese e poi la traduzione finale della risposta in italiano: **Domanda tradotta in inglese:** "At which university did Rubén Saúl Stiglitz obtain his Law degree?" **Risposta in inglese:** Rubén Saúl Stiglitz obtained his Law degree from the University of Buenos Aires. **Risposta finale:** Rubén Saúl Stiglitz ha conseguito la laurea in Giurisprudenza presso l'Università di Buenos Aires. \\
        \midrule
        \textbf{Chosen Path (Success)} & \textit{Raw Output:} La lingua più probabile in cui trovare informazioni per questa domanda è lo spagnolo, data l'origine ispanica del nome e la probabilità che una persona con questo nome abbia conseguito la laurea in un paese di lingua spagnola (in questo caso l'Argentina). **Domanda in spagnolo:** "¿En qué universidad Rubén Saúl Stiglitz obtuvo su título de abogado?" **Risposta in spagnolo:** Rubén Saúl Stiglitz obtuvo su título de abogado en la Universidad Nacional de La Plata. **Risposta finale:** Rubén Saúl Stiglitz ha conseguito la laurea in Giurisprudenza presso l'Università Nazionale di La Plata. \\
        \bottomrule
    \end{tabular}
    \caption{Example showing the model correctly routing to Spanish for an Argentine figure, whereas English reasoning hallucinates the most common Argentine university.}
\end{table*}

\begin{table*}
    \centering
    \small
    \begin{tabular}{p{0.15\linewidth} p{0.8\linewidth}}
        \toprule
        \textbf{Original Query} & Que ingrediente é adicionado ao Gloriao além de açúcar queimado, água fervida e aguardente? (Portuguese) \\
        \textbf{Gold Answer} & Café \\
        \midrule
        \textbf{English Path (Failed)} & \textit{Raw Output:} Aqui está a tradução da pergunta, a resposta em inglês e a tradução final em português: **Tradução da pergunta:** What ingredient is added to Gloriao besides burnt sugar, boiled water, and aguardente? **Resposta em inglês:** The ingredient added to Gloriao besides burnt sugar, boiled water, and aguardente is lemon zest. **Resposta final:** Raspa de limão. \\
        \midrule
        \textbf{Chosen Path (Success)} & \textit{Raw Output:} A pergunta refere-se a "Gloriao", que é uma bebida tradicional portuguesa. Portanto, é mais provável que as informações detalhadas... sejam encontradas em português. **Tradução da pergunta para português:** "Que ingrediente é adicionado ao Gloriao além de açúcar queimado, água fervida e aguardente?" **Pesquisa e Resposta em português:** O Gloriao é uma bebida alcoólica quente tipicamente portuguesa. A receita tradicional inclui açúcar queimado, água a ferver, aguardente e, muitas vezes, café ou casca de limão... o ingrediente mais comumente adicionado... é o café. **Resposta final:** Café \\
        \bottomrule
    \end{tabular}
    \caption{Example demonstrating access to highly localized cultural knowledge (a traditional Iberian drink) by reasoning in Portuguese instead of English.}
\end{table*}

\newpage

\section{Prompt Templates}
\label{sec:prompt_templates}

This section provides the full prompt templates used for each strategy across all supported languages. These instructions are always presented to the model in the original query language $L_q$.

\subsection*{Strategy: Native (x13) / Native Baseline}
\noindent\textbf{en:} \texttt{\small Answer the following factual question: '{question}'. Provide your final answer in English. The final answer should be concise and directly answer the question. Provide your final answer at the very end after the phrase 'Final Answer:'.} \\
\noindent\textbf{he:} \langHE{\texttt{\small ענה על השאלה העובדתית הבאה: '{question}'. ספק את התשובה הסופית בעברית. התשובה הסופית צריכה להיות תמציתית ולענות ישירות על השאלה. ספק את תשובתך הסופית בסוף לאחר הביטוי 'תשובה סופית:'.}} \\
\noindent\textbf{es:} \texttt{\small Responde a la siguiente pregunta factual: '{question}'. Proporciona tu respuesta final en español. La respuesta final debe ser concisa e responder directamente a la pregunta. Proporciona tu respuesta final al final después de la frase 'Respuesta final:'.} \\
\noindent\textbf{fr:} \texttt{\small Répondez à la question factuelle suivante : '{question}'. Fournissez votre réponse finale en français. La réponse finale doit être concise et répondre directement à la question. Fournissez votre réponse finale à la fin après la phrase 'Réponse finale:'.} \\
\noindent\textbf{ar:} \langAR{\texttt{\small أجب على السؤال الواقعي التالي: '{question}'. قدم إجابتك النهائية باللغة العربية. يجب أن تكون الإجابة النهائية موجزة وتجيب مباشرة على السؤال. قدم إجابتك النهائية في النهاية بعد العبارة 'الإجابة النهائية:'.}} \\
\noindent\textbf{ru:} \langRU{\texttt{\small Ответьте на следующий фактический вопрос: '{question}'. Предоставьте свой окончательный ответ на русском языке. Окончательный ответ должен быть кратким и прямо отвечать на вопрос. Предоставьте свой окончательный ответ в самом конце после фразы 'Окончательный ответ:'.}} \\
\noindent\textbf{it:} \texttt{\small Rispondi alla seguente domanda fattuale: '{question}'. Fornisci la tua risposta finale in italiano. La risposta finale dovrebbe essere concisa e rispondere direttamente alla domanda. Fornisci la tua risposta finale alla fine dopo la frase 'Risposta finale:'.} \\
\noindent\textbf{uk:} \langUK{\texttt{\small Дайте відповідь на наступне фактичне питання: '{question}'. Надайте свою остаточну відповідь українською. Остаточна відповідь повинна бути стислою та прямо відповідати на питання. Надайте свою остаточну відповідь в самому кінці після фрази 'Остаточна відповідь:'.}} \\
\noindent\textbf{bg:} \langBG{\texttt{\small Отговорете на следния фактически въпрос: '{question}'. Предоставете окончателния си отговор на български. Окончателният отговор трябва да бъде кратък и да отговаря директно на въпроса. Предоставете окончателния си отговор в самия край след фразата 'Окончателен отговор:'.}} \\
\noindent\textbf{zh:} \langZH{\texttt{\small 回答以下事实问题： '{question}'. 请用中文提供最终答案。 最终答案应简洁并直接回答问题。 在最后，请在以下短语后提供你的最终答案： '最终答案：'.}} \\
\noindent\textbf{hi:} \texttt{\small निम्नलिखित तथ्यात्मक प्रश्न का उत्तर दें: '{question}'. अपना अंतिम उत्तर हिंदी में दें। अंतिम उत्तर संक्षिप्त होना चाहिए और सीधे प्रश्न का उत्तर देना चाहिए। अंत में, इस वाक्यांश के बाद अपना अंतिम उत्तर दें 'अंतिम उत्तर:'.} \\
\noindent\textbf{bn:} \texttt{\small নিম্নলিখিত তথ্যগত প্রশ্নের উত্তর দিন: '{question}'. আপনার চূড়ান্ত উত্তর বাংলায় দিন। চূড়ান্ত উত্তরটি সংক্ষিপ্ত হওয়া উচিত এবং সরাসরি প্রশ্নের উত্তর দেওয়া উচিত। একেবারে শেষে, এই শব্দগুচ্ছের পরে আপনার চূড়ান্ত উত্তর দিন 'চূড়ান্ত উত্তর:'.} \\
\noindent\textbf{ja:} \langJA{\texttt{\small 次の事実に関する質問に答えてください： '{question}'. 最終的な回答を日本語で提供してください。 最終的な回答は簡潔で、質問に直接答える必要があります。 最後に、次のフレーズの後に最終回答を提供してください： '最終回答：'.}} \\
\noindent\textbf{id:} \texttt{\small Jawablah pertanyaan faktual berikut: '{question}'. Berikan jawaban akhir Anda dalam bahasa Indonesia. Jawaban akhir harus ringkas dan langsung menjawab pertanyaan. Berikan jawaban akhir Anda di bagian paling akhir setelah frasa 'Jawaban akhir:'.} \\
\noindent\textbf{ko:} \langKO{\texttt{\small 다음 사실 질문에 답하십시오: '{question}'. 최종 답변을 한국어로 제공하세요. 최종 답변은 간결하고 질문에 직접 답해야 합니다. 맨 마지막에 다음 문구 뒤에 최종 답변을 제공하십시오. '최종 답변:'.}} \\
\noindent\textbf{pt:} \texttt{\small Responda à seguinte pergunta factual: '{question}'. Forneça sua resposta final em português. A resposta final deve ser concisa e responder directamente à pergunta. Forneça sua resposta final no final após a frase 'Resposta final:'.} \\
\noindent\textbf{de:} \texttt{\small Beantworten Sie die folgende Sachfrage: '{question}'. Geben Sie Ihre endgültige Antwort auf Deutsch. Die endgültige Antwort sollte prägnant sein und die Frage direkt beantworten. Geben Sie Ihre endgültige Antwort ganz am Ende nach dem Satz an 'Endgültige Antwort:'.} \\

\subsection*{Strategy: $L_e \in \mathrm{Choose}$ (x13) / $L_e \rightarrow \mathrm{Choose\text{-}Unlimited}$}
\noindent\textbf{en:} \texttt{\small Choose the language where information for the following factual question is most likely to be found, translate the question to that language, answer it, and then translate the final answer back to English. The question is: '{question}'. Provide your final answer in English. The final answer should be concise and directly answer the question. Provide your final answer at the very end after the phrase 'Final Answer:'.} \\
\noindent\textbf{he:} \langHE{\texttt{\small בחר את השפה שבה הסיכוי הגבוה ביותר למצוא מידע עבור השאלה העובדתית הבאה, תרגם את השאלה לשפה זו, ענה עליה, ולאחר מכן תרגם את התשובה הסופית חזרה ל-עברית. השאלה היא: '{question}'. ספק את התשובה הסופית בעברית. התשובה הסופית צריכה להיות תמציתית ולענות ישירות על השאלה. ספק את תשובתך הסופית בסוף לאחר הביטוי 'תשובה סופית:'.}} \\
\noindent\textbf{es:} \texttt{\small Elige el idioma en el que es más probable encontrar información para la siguiente pregunta factual, traduce la pregunta a ese idioma, respóndela y luego traduce la respuesta final de vuelta al español. La pregunta es: '{question}'. Proporciona tu respuesta final en español. La respuesta final debe ser concisa e responder directamente a la pregunta. Proporciona tu respuesta final al final después de la frase 'Respuesta final:'.} \\
\noindent\textbf{fr:} \texttt{\small Choisissez la langue dans laquelle les informations pour la question factuelle suivante ont le plus de chances d'être trouvées, traduisez la question dans cette langue, répondez-y, puis traduisez la réponse finale de retour en français. La question est : '{question}'. Fournissez votre réponse finale en français. La réponse finale doit être concise et répondre directement à la question. Fournissez votre réponse finale à la fin après la phrase 'Réponse finale:'.} \\
\noindent\textbf{ar:} \langAR{\texttt{\small اختر اللغة التي يُحتمل أن تجد فيها معلومات للسؤال الواقعي التالي، ترجم السؤال إلى تلك اللغة، وأجب عليه، ثم ترجم الإجابة النهائية مرة أخرى إلى العربية. السؤال هو: '{question}'. قدم إجابتك النهائية باللغة العربية. يجب أن تكون الإجابة النهائية موجزة وتجيب مباشرة على السؤال. قدم إجابتك النهائية في النهاية بعد العبارة 'الإجابة النهائية:'.}} \\
\noindent\textbf{ru:} \langRU{\texttt{\small Выберите язык, на котором вероятнее всего будет найдена информация для следующего фактического вопроса, переведите вопрос на этот язык, ответьте на него, а затем переведите окончательный ответ обратно на русском. Вопрос: '{question}'. Предоставьте свой окончательный ответ на русском языке. Окончательный ответ должен быть кратким и прямо отвечать на вопрос. Предоставьте свой окончательный ответ в самом конце после фразы 'Окончательный ответ:'.}} \\
\noindent\textbf{it:} \texttt{\small Scegli la lingua in cui è più probabile trovare informazioni per la seguente domanda fattuale, traduci la domanda in quella lingua, rispondi e infine traduci la risposta finale di nuovo in italiano. La domanda è: '{question}'. Fornisci la tua risposta finale in italiano. La risposta finale dovrebbe essere concisa e rispondere direttamente alla domanda. Fornisci la tua risposta finale alla fine dopo la frase 'Risposta finale:'.} \\
\noindent\textbf{uk:} \langUK{\texttt{\small Виберіть мову, якою найімовірніше буде знайдено інформацію для наступного фактичного питання, перекладіть питання цією мовою, дайте відповідь на нього, а потім перекладіть остаточну відповідь назад на українською. Питання: '{question}'. Надайте свою остаточну відповідь українською. Остаточна відповідь повинна бути стислою та прямо відповідати на питання. Надайте свою остаточну відповідь в самому кінці після фрази 'Остаточна відповідь:'.}} \\
\noindent\textbf{bg:} \langBG{\texttt{\small Изберете езика, на който е най-вероятно да се намери информация за следния фактически въпрос, преведете въпроса на този език, отговорете на него и след това преведете окончателния отговор обратно на български. Въпросът е: '{question}'. Предоставете окончателния си отговор на български. Окончателният отговор трябва да бъде кратък и да отговаря директно на въпроса. Предоставете окончателния си отговор в самия край след фразата 'Окончателен отговор:'.}} \\
\noindent\textbf{zh:} \langZH{\texttt{\small 选择最有可能找到以下事实问题信息的语言，将问题翻译成该语言并回答，最后将最终答案翻译回中文。问题是： '{question}'. 请用中文提供最终答案。 最终答案应简洁并直接回答问题。 在最后，请在以下短语后提供你的最终答案： '最终答案：'.}} \\
\noindent\textbf{hi:} \texttt{\small वह भाषा चुनें जिसमें निम्नलिखित तथ्यात्मक प्रश्न की जानकारी मिलने की सबसे अधिक संभावना है, प्रश्न का उस भाषा में अनुवाद करें, उसका उत्तर दें, और फिर अंतिम उत्तर का हिंदी में वापस अनुवाद करें। प्रश्न है: '{question}'. अपना अंतिम उत्तर हिंदी में दें। अंतिम उत्तर संक्षिप्त होना चाहिए और सीधे प्रश्न का उत्तर देना चाहिए। अंत में, इस वाक्यांश के बाद अपना अंतिम उत्तर दें 'अंतिम उत्तर:'.} \\
\noindent\textbf{bn:} \texttt{\small এমন একটি ভাষা বেছে নিন যেখানে নিচের তথ্যগত প্রশ্নের তথ্য পাওয়ার সম্ভাবনা সবচেয়ে বেশি, প্রশ্নটি সেই ভাষায় অনুবাদ করুন, উত্তর দিন এবং তারপর চূড়ান্ত উত্তরটি বাংলা এ অনুবাদ করুন। প্রশ্নটি হলো: '{question}'. আপনার চূড়ান্ত উত্তর বাংলায় দিন। চূড়ান্ত উত্তরটি সংক্ষিপ্ত হওয়া উচিত এবং সরাসরি প্রশ্নের উত্তর দেওয়া উচিত। একেবারে শেষে, এই শব্দগুচ্ছের পরে আপনার চূড়ান্ত উত্তর দিন 'চূড়ান্ত উত্তর:'.} \\
\noindent\textbf{ja:} \langJA{\texttt{\small 次の事実に関する質問の情報が最も見つかりやすい言語を選択し、質問をその 言語に翻訳して答え、最終的な答えを日本語に翻訳し返してください。質問： '{question}'. 最終的な回答を日本語で提供してください。 最終的な回答は簡潔で、質問に直接答える必要があります。 最後に、次のフレーズの後に最終回答を提供してください： '最終回答：'.}} \\
\noindent\textbf{id:} \texttt{\small Pilih bahasa di mana informasi untuk pertanyaan faktual berikut paling mungkin ditemukan, terjemahkan pertanyaan tersebut ke bahasa itu, jawablah, dan kemudian terjemahkan jawaban akhirnya kembali ke Indonesia. Pertanyaannya adalah: '{question}'. Berikan jawaban akhir Anda dalam bahasa Indonesia. Jawaban akhir harus ringkas dan langsung menjawab pertanyaan. Berikan jawaban akhir Anda di bagian paling akhir setelah frasa 'Jawaban akhir:'.} \\
\noindent\textbf{ko:} \langKO{\texttt{\small 다음 사실적인 질문에 대한 정보를 찾을 가능성이 가장 높은 언어를 선택하고, 질문을 해당 언어로 번역하여 답한 다음, 최종 답변을 다시 한국어로 번역하십시오. 질문: '{question}'. 최종 답변을 한국어로 제공하세요. 최종 답변은 간결하고 질문에 직접 답해야 합니다. 맨 마지막에 다음 문구 뒤에 최종 답변을 제공하십시오. '최종 답변:'.}} \\
\noindent\textbf{pt:} \texttt{\small Escolha o idioma no qual é mais provável que as informações para a seguinte pergunta factual sejam encontradas, traduza a pergunta para esse idioma, responda a ela e, em seguida, traduza a resposta final de volta para o português. A pergunta é: '{question}'. Forneça sua resposta final em português. A resposta final deve ser concisa e responder directamente à pergunta. Forneça sua resposta final no final após a frase 'Resposta final:'.} \\
\noindent\textbf{de:} \texttt{\small Wählen Sie die Sprache aus, in der am wahrscheinlichsten Informationen zu der folgenden Sachfrage gefunden werden, übersetzen Sie die Frage in diese Sprache, beantworten Sie sie und übersetzen Sie die endgültige Antwort dann zurück ins Deutsch. Die Frage lautet: '{question}'. Geben Sie Ihre endgültige Antwort auf Deutsch. Die endgültige Antwort sollte prägnant sein und die Frage direkt beantworten. Geben Sie Ihre endgültige Antwort ganz am Ende nach dem Satz an 'Endgültige Antwort:'.} \\

\subsection*{Strategy: $L_e \rightarrow \mathrm{Choose\text{-}ReasonAboutOrigin}$}
\noindent\textbf{en:} \texttt{\small Keep in mind that the knowledge for the following factual question might not be in English or other popular languages. First, think about where this knowledge most likely originates and which language would be most appropriate to find it. Then, choose that language, translate the question to it, reason about it step-by-step in that language, answer it, and finally translate the final answer back to English. The question is: '{question}'. Provide your final answer in English. The final answer should be concise and directly answer the question. Provide your final answer at the very end after the phrase 'Final Answer:'.} \\
\noindent\textbf{he:} \langHE{\texttt{\small זכור שהידע עבור השאלה העובדתית הבאה עשוי שלא להימצא באנגלית או בשפות נפוצות אחרות. ראשית, חשוב היכן הידע הזה ככל הנראה מקורו ואיזו שפה תהיה המתאימה ביותר למציאתו. לאחר מכן, בחר את השפה הזו, תרגם את השאלה אליה, חשוב עליה צעד אחר צעד באותה שפה, ענה עליה, ולבסוף תרגם את התשובה הסופית חזרה ל-עברית. השאלה היא: '{question}'. ספק את התשובה הסופית בעברית. התשובה הסופית צריכה להיות תמציתית ולענות ישירות על השאלה. ספק את תשובתך הסופית בסוף לאחר הביטוי 'תשובה סופית:'.}} \\
\noindent\textbf{es:} \texttt{\small Ten en cuenta que el conocimiento para la siguiente pregunta factual podría no estar en inglés u otros idiomas populares. Primero, piensa en dónde se origina más probablemente este conocimiento y qué idioma sería el más apropiado para encontrarlo. Luego, elige ese idioma, traduce la pregunta a él, razona sobre ella paso a paso en ese idioma, respóndela y finalmente traduce la respuesta final de vuelta al español. La pregunta es: '{question}'. Proporciona tu respuesta final en español. La respuesta final debe ser concisa e responder directamente a la pregunta. Proporciona tu respuesta final al final después de la frase 'Respuesta final:'.} \\
\noindent\textbf{fr:} \texttt{\small Gardez à l'esprit que les connaissances pour la question factuelle suivante pourraient ne pas être en anglais ou dans d'autres langues populaires. Tout d'abord, réfléchissez à l'origine la plus probable de ces connaissances et à la langue la plus appropriée pour les trouver. Ensuite, choisissez cette langue, traduisez la question dans celle-ci, raisonnez à ce sujet étape par étape dans cette langue, répondez-y et enfin traduisez la réponse finale de retour en français. La question est : '{question}'. Fournissez votre réponse finale en français. La réponse finale doit être concise et répondre directement à la question. Fournissez votre réponse finale à la fin après la phrase 'Réponse finale:'.} \\
\noindent\textbf{ar:} \langAR{\texttt{\small ضع في اعتبارك أن المعرفة الخاصة بالسؤال الواقعي التالي قد لا تكون باللغة الإنجليزية أو غيرها من اللغات الشائعة. أولاً، فكر في المكان الذي من المرجح أن تنشأ فيه هذه المعرفة واللغة التي ستكون الأنسب للعثور عليها. ثم، اختر تلك اللغة، وترجم السؤال إليها، وفكر فيه خطوة بخطوة بتلك اللغة، وأجب عليه، وأخيرًا ترجم الإجابة النهائية مرة أخرى إلى العربية. السؤال هو: '{question}'. قدم إجابتك النهائية باللغة العربية. يجب أن تكون الإجابة النهائية موجزة وتجيب مباشرة على السؤال. قدم إجابتك النهائية في النهاية بعد العبارة 'الإجابة النهائية:'.}} \\
\noindent\textbf{ru:} \langRU{\texttt{\small Имейте в виду, что знания для следующего фактического вопроса могут быть не на английском или других популярных языках. Сначала подумайте о том, где вероятнее всего зародились эти знания и какой язык был бы наиболее подходящим для их поиска. Затем выберите этот язык, переведите на него вопрос, обдумайте его шаг за шагом на этом языке, ответьте на него и, наконец, переведите окончательный ответ обратно на русском. Вопрос: '{question}'. Предоставьте свой окончательный ответ на русском языке. Окончательный ответ должен быть кратким и прямо отвечать на вопрос. Предоставьте свой окончательный ответ в самом конце после фразы 'Окончательный ответ:'.}} \\
\noindent\textbf{it:} \texttt{\small Tieni presente che la conoscenza per la seguente domanda fattuale potrebbe non essere in inglese o in altre lingue popolari. Per prima cosa, pensa a dove ha più probabilmente origine questa conoscenza e quale lingua sarebbe la più appropriata per trovarla. Quindi, scegli quella lingua, traduci la domanda in essa, ragiona su di essa passo dopo passo in quella lingua, rispondi e infine traduci la risposta finale di nuovo in italiano. La domanda è: '{question}'. Fornisci la tua risposta finale in italiano. La risposta finale dovrebbe essere concisa e rispondere direttamente alla domanda. Fornisci la tua risposta finale alla fine dopo la frase 'Risposta finale:'.} \\
\noindent\textbf{uk:} \langUK{\texttt{\small Майте на увазі, що знання для наступного фактичного питання можуть бути не англійською або іншими популярними мовами. Спочатку подумайте про те, де найімовірніше зародилися ці знання і яка мова була б найбільш підходящою для їх пошуку. Потім виберіть цю мову, перекладіть нею питання, обміркуйте його крок за кроком цією мовою, дайте відповідь на нього і, нарешті, перекладіть остаточну відповідь назад на українською. Питання: '{question}'. Надайте свою остаточну відповідь українською. Остаточна відповідь повинна бути стислою та прямо відповідати на питання. Надайте свою остаточну відповідь в самому кінці після фрази 'Остаточна відповідь:'.}} \\
\noindent\textbf{bg:} \langBG{\texttt{\small Имайте предвид, че знанията за следния фактически въпрос може да не са на английски или други популярни езици. Първо, помислете къде най-вероятно възникват тези знания и кой език би бил най-подходящ за намирането им. След това изберете този език, преведете въпроса на него, разсъждавайте върху него стъпка по стъпка на този език, отговорете му и накрая преведете окончателния отговор обратно на български. Въпросът е: '{question}'. Предоставете окончателния си отговор на български. Окончателният отговор трябва да бъде кратък и да отговаря директно на въпроса. Предоставете окончателния си отговор в самия край след фразата 'Окончателен отговор:'.}} \\
\noindent\textbf{zh:} \langZH{\texttt{\small 请记住，以下事实问题的知识可能不是英语或其他流行语言。首先，思考这些知识最有可能起源于哪里，以及哪种语言最适合查找它。然后，选择那种语言，将问题翻译成该语言，用该语言逐步推理，回答问题，最后将最终答案翻译回中文。问题是： '{question}'. 请用中文提供最终答案。 最终答案应简洁并直接回答问题。 在最后，请在以下短语后提供你的最终答案： '最终答案：'.}} \\
\noindent\textbf{hi:} \texttt{\small ध्यान रखें कि निम्नलिखित तथ्यात्मक प्रश्न का ज्ञान अंग्रेजी या अन्य लोकप्रिय भाषाओं में नहीं हो सकता है। सबसे पहले, सोचें कि यह ज्ञान सबसे अधिक कहाँ से उत्पन्न होता है और इसे खोजने के लिए कौन सी भाषा सबसे उपयुक्त होगी। फिर, वह भाषा चुनें, प्रश्न का उसमें अनुवाद करें, उस भाषा में चरण-दर-चरण इसके बारे में तर्क करें, इसका उत्तर दें, और अंत में अंतिम उत्तर का हिंदी में वापस अनुवाद करें। प्रश्न है: '{question}'. अपना अंतिम उत्तर हिंदी में दें। अंतिम उत्तर संक्षिप्त होना चाहिए और सीधे प्रश्न का उत्तर देना चाहिए। अंत में, इस वाक्यांश के बाद अपना अंतिम उत्तर दें 'अंतिम उत्तर:'.} \\
\noindent\textbf{bn:} \texttt{\small মনে রাখবেন যে নিচের তথ্যগত প্রশ্নের জ্ঞান ইংরেজি বা অন্যান্য জনপ্রিয় ভাষায় নাও থাকতে পারে। প্রথমত, চিন্তা করুন এই জ্ঞানের উৎপত্তি সম্ভবত কোথায় এবং এটি খুঁজে পেতে কোন ভাষা সবচেয়ে উপযুক্ত হবে। তারপরে, সেই ভাষাটি বেছে নিন, প্রশ্নটি সেই ভাষায় অনুবাদ করুন, সেই ভাষায় ধাপে ধাপে এটি নিয়ে যুক্তি করুন, উত্তর দিন এবং অবশেষে চূড়ান্ত উত্তরটি বাংলা-এ অনুবাদ করুন। প্রশ্নটি হলো: '{question}'. আপনার চূড়ান্ত উত্তর বাংলায় দিন। চূড়ান্ত উত্তরটি সংক্ষিপ্ত হওয়া উচিত এবং সরাসরি প্রশ্নের উত্তর দেওয়া উচিত। একেবারে শেষে, এই শব্দগুচ্ছের পরে আপনার চূড়ান্ত উত্তর দিন 'চূড়ান্ত উত্তর:'.} \\
\noindent\textbf{ja:} \langJA{\texttt{\small 次の事実に関する質問の知識は、英語やその他の一般的な言語ではない可能性があることに留意してください。まず、この知識が最も発生しそうな場所と、それを見つけるのに最も適した言語について考えてください。次に、その言語を選択し、質問をその言語に翻訳し、その言語で段階的に推論し、回答し、最後に最終的な回答を日本語に翻訳し返してください。質問： '{question}'. 最終的な回答を日本語で提供してください。 最終的な回答は簡潔で、質問に直接答える必要があります。 最後に、次のフレーズの後に最終回答を提供してください： '最終回答：'.}} \\
\noindent\textbf{id:} \texttt{\small Ingatlah bahwa pengetahuan untuk pertanyaan faktual berikut ini mungkin tidak dalam bahasa Inggris atau bahasa populer lainnya. Pertama, pikirkan dari mana pengetahuan ini kemungkinan besar berasal dan bahasa apa yang paling tepat untuk menemukannya. Kemudian, pilih bahasa tersebut, terjemahkan pertanyaannya ke bahasa itu, pikirkan secara bertahap dalam bahasa itu, jawablah, dan terakhir terjemahkan jawaban akhirnya kembali ke Indonesia. Pertanyaannya adalah: '{question}'. Berikan jawaban akhir Anda dalam bahasa Indonesia. Jawaban akhir harus ringkas dan langsung menjawab pertanyaan. Berikan jawaban akhir Anda di bagian paling akhir setelah frasa 'Jawaban akhir:'.} \\
\noindent\textbf{ko:} \langKO{\texttt{\small 다음 사실적인 질문에 대한 지식은 영어나 다른 대중적인 언어에 없을 수도 있음을 명심하십시오. 먼저 이 지식이 가장 기원했을 가능성이 높은 곳과 이를 찾기에 가장 적절한 언어가 무엇인지 생각해 보십시오. 그런 다음 해당 언어를 선택하고 질문을 해당 언어로 번역하고 해당 언어로 단계별로 추론하여 답한 다음 마지막으로 최종 답변을 한국어로 다시 번역하십시오. 질문: '{question}'. 최종 답변을 한국어로 제공하세요. 최종 답변은 간결하고 질문에 직접 답해야 합니다. 맨 마지막에 다음 문구 뒤에 최종 답변을 제공하십시오. '최종 답변:'.}} \\
\noindent\textbf{pt:} \texttt{\small Tenha em mente que o conhecimento para a seguinte pergunta factual pode não estar em inglês ou outros idiomas populares. Primeiro, pense sobre onde esse conhecimento mais provavelmente se origina e qual idioma seria o mais apropriado para encontrá-lo. Então, escolha esse idioma, traduza a pergunta para ele, raciocine sobre ela passo a passo nesse idioma, responda a ela e, finalmente, traduza a resposta final de volta para o português. A pergunta é: '{question}'. Forneça sua resposta final em português. A resposta final deve ser concisa e responder directamente à pergunta. Forneça sua resposta final no final após a frase 'Resposta final:'.} \\
\noindent\textbf{de:} \texttt{\small Beachten Sie, dass das Wissen für die folgende Sachfrage möglicherweise nicht auf Englisch oder in anderen populären Sprachen vorliegt. Überlegen Sie zunächst, woher dieses Wissen am wahrscheinlichsten stammt und welche Sprache am besten geeignet wäre, um es zu finden. Wählen Sie dann diese Sprache aus, übersetzen Sie die Frage in diese Sprache, denken Sie Schritt für Schritt in dieser Sprache darüber nach, beantworten Sie sie und übersetzen Sie schließlich die endgültige Antwort zurück ins Deutsch. Die Frage lautet: '{question}'. Geben Sie Ihre endgültige Antwort auf Deutsch. Die endgültige Antwort sollte prägnant sein und die Frage direkt beantworten. Geben Sie Ihre endgültige Antwort ganz am Ende nach dem Satz an 'Endgültige Antwort:'.} \\

\subsection*{Strategy: $L_e \rightarrow \mathrm{Choose} + \mathrm{EN}$}
\noindent\textbf{en:} \texttt{\small In addition to English, choose one more language where information for the following factual question is most likely to be found. Translate the question to both English and that language, answer it in both, and then translate the final answer back to English. The question is: '{question}'. Provide your final answer in English. The final answer should be concise and directly answer the question. Provide your final answer at the very end after the phrase 'Final Answer:'.} \\
\noindent\textbf{he:} \langHE{\texttt{\small בנוסף לאנגלית, בחר שפה אחת נוספת שבה הסיכוי הגבוה ביותר למצוא מידע עבור השאלה העובדתית הבאה. תרגם את השאלה גם לאנגלית וגם לשפה שבחרת, ענה עליה בשתי השפות, ולאחר מכן תרגם את התשובה הסופית חזרה ל-עברית. השאלה היא: '{question}'. ספק את התשובה הסופית בעברית. התשובה הסופית צריכה להיות תמציתית ולענות ישירות על השאלה. ספק את תשובתך הסופית בסוף לאחר הביטוי 'תשובה סופית:'.}} \\
\noindent\textbf{es:} \texttt{\small Además del inglés, elige un idioma más en el que sea más probable encontrar información para la siguiente pregunta factual. Traduce la pregunta tanto al inglés como a ese idioma, respóndela en ambos y luego traduce la respuesta final de vuelta al español. La pregunta es: '{question}'. Proporciona tu respuesta final en español. La respuesta final debe ser concisa e responder directamente a la pregunta. Proporciona tu respuesta final al final después de la frase 'Respuesta final:'.} \\
\noindent\textbf{fr:} \texttt{\small En plus de l'anglais, choisissez une autre langue dans laquelle les informations pour la question factuelle suivante ont le plus de chances d'être trouvées. Traduisez la question en anglais et dans cette langue, répondez-y dans les deux langues, puis traduisez la réponse finale en français. La question est : '{question}'. Fournissez votre réponse finale en français. La réponse finale doit être concise et répondre directement à la question. Fournissez votre réponse finale à la fin après la phrase 'Réponse finale:'.} \\
\noindent\textbf{ar:} \langAR{\texttt{\small بالإضافة إلى اللغة الإنجليزية، اختر لغة أخرى يُحتمل أن تجد فيها معلومات للسؤال الواقعי التالي. ترجم السؤال إلى كل من اللغة الإنجليزية وتلك اللغة، ואجب عليه بكلتا اللגتين، ثم ترجم الإגابة النهائية مرة أخرى إلى العربية. السؤال هو: '{question}'. قدم إجابتك النهائية باللغة العربية. يجب أن تكون الإجابة النهائية موجزة وتجيب مباشرة على السؤال. قدم إجابتك النهائية في النهاية بعد العبارة 'الإجابة النهائية:'.}} \\
\noindent\textbf{ru:} \langRU{\texttt{\small В дополнение к английскому языку выберите еще один язык, на котором вероятнее всего будет найдена информация для следующего фактического вопроса. Переведите вопрос на английский и на этот язык, ответьте на него на обоих языках, а затем переведите окончательный ответ обратно на русском. Вопрос: '{question}'. Предоставьте свой окончательный ответ на русском языке. Окончательный ответ должен быть кратким и прямо отвечать на вопрос. Предоставьте свой окончательный ответ в самом конце после фразы 'Окончательный ответ:'.}} \\
\noindent\textbf{it:} \texttt{\small Oltre all'inglese, scegli un'altra lingua in cui è più probabile trovare informazioni per la seguente domanda fattuale. Traduci la domanda sia in inglese che in quella lingua, rispondi in entrambe le lingue e infine traduci la risposta finale di nuovo in italiano. La domanda è: '{question}'. Fornisci la tua risposta finale in italiano. La risposta finale dovrebbe essere concisa e rispondere direttamente alla domanda. Fornisci la tua risposta finale alla fine dopo la frase 'Risposta finale:'.} \\
\noindent\textbf{uk:} \langUK{\texttt{\small На додаток до англійської, виберіть ще одну мову, якою найімовірніше буде знайдено інформацію для наступного фактичного питання. Перекладіть питання англійською та цією мовою, дайте відповідь на нього обома мовами, а потім перекладіть остаточну відповідь назад на українською. Питання: '{question}'. Надайте свою остаточну відповідь українською. Остаточна відповідь повинна бути стислою та прямо відповідати на питання. Надайте свою остаточну відповідь в самому кінці після фрази 'Остаточна відповідь:'.}} \\
\noindent\textbf{bg:} \langBG{\texttt{\small В допълнение към английския език, изберете още един език, на който е най-вероятно да се намери информация за следния фактически въпрос. Преведете въпроса на английски и на този език, отговорете на него и на двата езика и след това преведете окончателния отговор обратно на български. Въпросът е: '{question}'. Предоставете окончателния си отговор на български. Окончателният отговор трябва да бъде кратък и да отговаря директно на въпроса. Предоставете окончателния си отговор в самия край след фразата 'Окончателен отговор:'.}} \\
\noindent\textbf{zh:} \langZH{\texttt{\small 除了英语之外，再选择一种最有可能找到以下事实问题信息的语言。将问题翻译成英语和该语言，用这两种语言回答，最后将最终答案翻译回中文。问题是： '{question}'. 请用中文提供最终答案。 最终答案应简洁并直接回答问题。 在最后，请在以下短语后提供你的最终答案： '最终答案：'.}} \\
\noindent\textbf{hi:} \texttt{\small अंग्रेजी के अलावा, एक और भाषा चुनें जिसमें निम्नलिखित तथ्यात्मक प्रश्न की जानकारी मिलने की सबसे अधिक संभावना है। प्रश्न का अंग्रेजी और उस भाषा दोनों में अनुवाद करें, दोनों में इसका उत्तर दें, और फिर अंतिम उत्तर का हिंदी में वापस अनुवाद करें। प्रश्न है: '{question}'. अपना अंतिम उत्तर हिंदी में दें। अंतिम उत्तर संक्षिप्त होना चाहिए और सीधे प्रश्न का उत्तर देना चाहिए। अंत में, इस वाक्यांश के बाद अपना अंतिम उत्तर दें 'अंतिम उत्तर:'.} \\
\noindent\textbf{bn:} \texttt{\small ইংরেজি ছাড়াও আরও একটি ভাষা বেছে নিন যেখানে নিচের তথ্যগত প্রশ্নের তথ্য পাওয়ার সম্ভাবনা סבבת সবচেয়ে বেশি। প্রশ্নটি ইংরেজি এবং সেই ভাষা উভয়ই অনুবাদ করুন, উভয় ভাষায় উত্তর দিন এবং তারপর চূড়ান্ত উত্তরটি বাংলা এ অনুবাদ করুন। প্রশ্নটি হলো: '{question}'. আপনার চূড়ান্ত উত্তর বাংলায় দিন। চূড়ান্ত উত্তরটি সংক্ষিপ্ত হওয়া উচিত এবং সরাসরি প্রশ্নের উত্তর দেওয়া উচিত। একেবারে শেষে, এই শব্দগুচ্ছের পরে আপনার চূড়ান্ত উত্তর দিন 'চূড়ান্ত উত্তর:'.} \\
\noindent\textbf{ja:} \langJA{\texttt{\small 英語に加えて、次の事実に関する質問の情報が最も見つかりやすい言語をもう1つ選択してください。質問を英語とその言語の両方に翻訳して両方で答え、最終的な答えを日本語に翻訳し返してください。質問： '{question}'. 最終的な回答を日本語で提供してください。 最終的な回答は簡潔で、質問に直接答える必要があります。 最後に、次のフレーズの後に最終回答を提供してください： '最終回答：'.}} \\
\noindent\textbf{id:} \texttt{\small Selain bahasa Inggris, pilih satu bahasa lagi di mana informasi untuk pertanyaan faktual berikut paling mungkin ditemukan. Terjemahkan pertanyaan tersebut ke dalam bahasa Inggris dan bahasa itu, jawablah dalam kedua bahasa, dan kemudian terjemahkan jawaban akhirnya kembali ke Indonesia. Pertanyaannya adalah: '{question}'. Berikan jawaban akhir Anda dalam bahasa Indonesia. Jawaban akhir harus ringkas dan langsung menjawab pertanyaan. Berikan jawaban akhir Anda di bagian paling akhir setelah frasa 'Jawaban akhir:'.} \\
\noindent\textbf{ko:} \langKO{\texttt{\small 영어 외에도 다음 사실적인 질문에 대한 정보를 찾을 가능성이 가장 높은 언어를 하나 더 선택하십시오. 질문을 영어와 해당 언어 모두로 번역하고, 두 언어 모두로 답한 다음, 최종 답변을 다시 한국어로 번역하십시오. 질문: '{question}'. 최종 답변을 한국어로 제공하세요. 최종 답변은 간결하고 질문에 직접 답해야 합니다. 맨 마지막에 다음 문구 뒤에 최종 답변을 제공하십시오. '최종 답변:'.}} \\
\noindent\textbf{pt:} \texttt{\small Além do inglês, escolha mais um idioma no qual é mais provável que as informações para a seguinte pergunta factual sejam encontradas. Traduza a pergunta tanto para o inglês quanto para esse idioma, responda em ambos e, em seguida, traduza a resposta final de volta para o português. A pergunta é: '{question}'. Forneça sua resposta final em português. A resposta final deve ser concisa e responder directamente à pergunta. Forneça sua resposta final no final após a frase 'Resposta final:'.} \\
\noindent\textbf{de:} \texttt{\small Wählen Sie neben Englisch eine weitere Sprache aus, in der am wahrscheinlichsten Informationen zu der folgenden Sachfrage gefunden werden. Übersetzen Sie die Frage sowohl ins Englische als auch in diese Sprache, beantworten Sie sie in beiden Sprachen und übersetzen Sie die endgültige Antwort dann zurück ins Deutsch. Die Frage lautet: '{question}'. Geben Sie Ihre endgültige Antwort auf Deutsch. Die endgültige Antwort sollte prägnant sein und die Frage direkt beantworten. Geben Sie Ihre endgültige Antwort ganz am Ende nach dem Satz an 'Endgültige Antwort:'.} \\

\subsection*{Strategy: $L_e \in \mathcal{L}$ / $L_e \in \mathcal{L}_{\mathrm{pred}}$ / $L_e \rightarrow \mathrm{EN}$} 

\noindent\textbf{en:} \texttt{\small Translate the following factual question to [REASONING\_LANGUAGE], then answer it, and then translate the final answer back to English. The question is: '{question}'. Provide your final answer in English. The final answer should be concise and directly answer the question. Provide your final answer at the very end after the phrase 'Final Answer:'.} \\
\noindent\textbf{he:} \langHE{\texttt{\small תרגם את השאלה העובדתית הבאה ל-[REASONING\_LANGUAGE], ואז ענה עליה, ולאחר מכן תרגם את התשובה הסופית חזרה ל-עברית. השאלה היא: '{question}'. ספק את התשובה הסופית בעברית. התשובה הסופית צריכה להיות תמציתית ולענות ישירות על השאלה. ספק את תשובתך הסופית בסוף לאחר הביטוי 'תשובה סופית:'.}} \\
\noindent\textbf{es:} \texttt{\small Traduce la siguiente pregunta factual al [REASONING\_LANGUAGE], luego respóndela, y después traduce la respuesta final de vuelta al español. La pregunta es: '{question}'. Proporciona tu respuesta final en español. La respuesta final debe ser concisa e responder directamente a la pregunta. Proporciona tu respuesta final al final después de la frase 'Respuesta final:'.} \\
\noindent\textbf{fr:} \texttt{\small Traduisez la question factuelle suivante en [REASONING\_LANGUAGE], puis répondez-y, et ensuite traduisez la réponse finale de retour en français. La question est : '{question}'. Fournissez votre réponse finale en français. La réponse finale doit être concise et répondre directement à la question. Fournissez votre réponse finale à la fin après la phrase 'Réponse finale:'.} \\
\noindent\textbf{ar:} \langAR{\texttt{\small ترجم السؤال الواقعي التالي إلى [REASONING\_LANGUAGE]، ثم أجب عليه، ثم ترجم الإجابة النهائية مرة أخرى إلى العربية. السؤال هو: '{question}'. قدم إجابتك النهائية باللغة العربية. يجب أن تكون الإجابة النهائية موجزة وتجيب مباشرة على السؤال. قدم إجابتك النهائية في النهاية بعد العبارة 'الإجابة النهائية:'.}} \\
\noindent\textbf{ru:} \langRU{\texttt{\small Переведите следующий фактический вопрос на [REASONING\_LANGUAGE], затем ответьте на него, а затем переведите окончательный ответ обратно на русском. Вопрос: '{question}'. Предоставьте свой окончательный ответ на русском языке. Окончательный ответ должен быть кратким и прямо отвечать на вопрос. Предоставьте свой окончательный ответ в самом конце после фразы 'Окончательный ответ:'.}} \\
\noindent\textbf{it:} \texttt{\small Traduci la seguente domanda fattuale in [REASONING\_LANGUAGE], poi rispondi, e infine traduci la risposta finale di nuovo in italiano. La domanda è: '{question}'. Fornisci la tua risposta finale in italiano. La risposta finale dovrebbe essere concisa e rispondere direttamente alla domanda. Fornisci la tua risposta finale alla fine dopo la frase 'Risposta finale:'.} \\
\noindent\textbf{uk:} \langUK{\texttt{\small Перекладіть наступне фактичне питання на [REASONING\_LANGUAGE], потім дайте відповідь на нього, а потім перекладіть остаточну відповідь назад на українською. Питання: '{question}'. Надайте свою остаточну відповідь українською. Остаточна відповідь повинна бути стислою та прямо відповідати на питання. Надайте свою остаточну відповідь в самому кінці після фрази 'Остаточна відповідь:'.}} \\
\noindent\textbf{bg:} \langBG{\texttt{\small Преведете следния фактически въпрос на [REASONING\_LANGUAGE], след това отговорете на него и след това преведете окончателния отговор обратно на български. Въпросът е: '{question}'. Предоставете окончателния си отговор на български. Окончателният отговор трябва да бъде кратък и да отговаря директно на въпроса. Предоставете окончателния си отговор в самия край след фразата 'Окончателен отговор:'.}} \\
\noindent\textbf{zh:} \langZH{\texttt{\small 将以下事实问题翻译成[REASONING\_LANGUAGE]，然后回答，最后将最终答案翻译回中文。问题是： '{question}'. 请用中文提供最终答案。 最终答案应简洁并直接回答问题。 在最后，请在以下短语后提供你的最终答案： '最终答案：'.}} \\
\noindent\textbf{hi:} \texttt{\small निम्नलिखित तथ्यात्मक प्रश्न का [REASONING\_LANGUAGE] में अनुवाद करें, फिर उसका उत्तर दें, और फिर अंतिम उत्तर का हिंदी में वापस अनुवाद करें। प्रश्न है: '{question}'. अपना अंतिम उत्तर हिंदी में दें। अंतिम उत्तर संक्षिप्त होना चाहिए और सीधे प्रश्न का उत्तर देना चाहिए। अंत में, इस वाक्यांश के बाद अपना अंतिम उत्तर दें 'अंतिम उत्तर:'.} \\
\noindent\textbf{bn:} \texttt{\small নিচের তথ্যগত প্রশ্নটি [REASONING\_LANGUAGE] এ অনুবাদ করুন, তারপর উত্তর দিন, এবং তারপর চূড়ান্ত উত্তরটি বাংলা এ অনুবাদ করুন। প্রশ্নটি হলো: '{question}'. আপনার চূড়ান্ত উত্তর বাংলায় দিন। চূড়ান্ত উত্তরটি সংক্ষিপ্ত হওয়া উচিত এবং সরাসরি প্রশ্নের উত্তর দেওয়া উচিত। একেবারে শেষে, এই শব্দগুচ্ছের পরে আপনার চূড়ান্ত উত্তর দিন 'চূড়ান্ত উত্তর:'.} \\
\noindent\textbf{ja:} \langJA{\texttt{\small 次の事実に関する質問を[REASONING\_LANGUAGE]に翻訳し、それに答えてから、最終的な答えを日本語に翻訳し返してください。質問： '{question}'. 最終的な回答を日本語で提供してください。 最終的な回答は簡潔で、質問に直接答える必要があります。 最後に、次のフレーズの後に最終回答を提供してください： '最終回答：'.}} \\
\noindent\textbf{id:} \texttt{\small Terjemahkan pertanyaan faktual berikut ke dalam [REASONING\_LANGUAGE], kemudian jawablah, dan kemudian terjemahkan jawaban akhir kembali ke Indonesia. Pertanyaannya adalah: '{question}'. Berikan jawaban akhir Anda dalam bahasa Indonesia. Jawaban akhir harus ringkas dan langsung menjawab pertanyaan. Berikan jawaban akhir Anda di bagian paling akhir setelah frasa 'Jawaban akhir:'.} \\
\noindent\textbf{ko:} \langKO{\texttt{\small 다음 사실 질문을 [REASONING\_LANGUAGE] (으)로 번역한 다음 답변하고, 최종 답변을 한국어 (으)로 다시 번역하십시오. 질문은 다음과 같습니다: '{question}'. 최종 답변을 한국어로 제공하세요. 최종 답변은 간결하고 질문에 직접 답해야 합니다. 맨 마지막에 다음 문구 뒤에 최종 답변을 제공하십시오. '최종 답변:'.}} \\
\noindent\textbf{pt:} \texttt{\small Traduza a seguinte pergunta factual para [REASONING\_LANGUAGE], em seguida, responda-a e depois traduza a resposta final de volta para português. A pergunta é: '{question}'. Forneça sua resposta final em português. A resposta final deve ser concisa e responder directamente à pergunta. Forneça sua resposta final no final após a frase 'Resposta final:'.} \\
\noindent\textbf{de:} \texttt{\small Übersetzen Sie die folgende Sachfrage in [REASONING\_LANGUAGE], beantworten Sie sie dann und übersetzen Sie schließlich die endgültige Antwort zurück in Deutsch. Die Frage lautet: '{question}'. Geben Sie Ihre endgültige Antwort auf Deutsch. Die endgültige Antwort sollte prägnant sein und die Frage direkt beantworten. Geben Sie Ihre endgültige Antwort ganz am Ende nach dem Satz an 'Endgültige Antwort:'.} \\

\subsection*{Strategy: $L_e \in \mathcal{L}_{\mathrm{pred}}$}
\noindent\textbf{en:} \texttt{\small Predict 3 languages that would be most effective for reasoning and finding the correct answer to the following factual question. Provide the response as a JSON object with a single key "languages" containing a list of ISO 639-1 language codes. Question: {question}} \\
\noindent\textbf{he:} \langHE{\texttt{\small חזה 3 שפות שיהיו היעילות ביותר עבור חשיבה ומציאת התשובה הנכונה לשאלה העובדתית הבאה. ספק את התשובה כאובייקט JSON עם מפתח יחיד "languages" המכיל רשימה של קודי ISO 639-1. שאלה: {question}}} \\
\noindent\textbf{es:} \texttt{\small Predice 3 idiomas que serían más efectivos para razonar y encontrar la respuesta correcta a la siguiente pregunta fáctica. Proporciona la respuesta como un objeto JSON con una única clave "languages" que contenga una lista de códigos de idioma ISO 639-1. Pregunta: {question}} \\
\noindent\textbf{fr:} \texttt{\small Prédisez 3 langues qui seraient les plus efficaces pour raisonner et trouver la bonne réponse à la question factuelle suivante. Fournissez la réponse sous la forme d'un objet JSON avec une seule clé "languages" contenant une liste de codes de langue ISO 639-1. Question : {question}} \\
\noindent\textbf{ar:} \langAR{\texttt{\small تنبأ بـ 3 لغات ستكون الأكثر فعالية للتفكير وإيجاد الإجابة الصحيحة للسؤال الواقعي التالي. قدم الإجابة ككائن JSON بمفتاح واحد "languages" يحتوي على قائمة برموز اللغات ISO 639-1. السؤال: {question}}} \\
\noindent\textbf{ru:} \langRU{\texttt{\small Предскажите 3 языков, которые были бы наиболее эффективны для рассуждений и поиска правильного ответа на следующий фактический вопрос. Предоставьте ответ в виде JSON-объекта с единственным ключом "languages", содержащим список кодов языков ISO 639-1. Вопрос: {question}}} \\
\noindent\textbf{it:} \texttt{\small Prevedi 3 lingue che sarebbero più efficaci per ragionare e trovare la risposta corretta alla seguente domanda fattuale. Fornisci la risposta come oggetto JSON con una singola chiave "languages" contenente un elenco di codici lingua ISO 639-1. Domanda: {question}} \\
\noindent\textbf{uk:} \langUK{\texttt{\small Передбачте 3 мов, які були б найбільш ефективними для міркувань та пошуку правильної відповіді на наступне фактичне питання. Надайте відповідь у вигляді JSON-об'єкта з єдиним ключем "languages", що містить список кодів мов ISO 639-1. Питання: {question}}} \\
\noindent\textbf{bg:} \langBG{\texttt{\small Предвидете 3 езика, които биха били най-ефективни за разсъждение и намиране на правилния отговор на следния фактически въпрос. Предоставете отговора като JSON обект с един ключ "languages", съдържащ списък с езикови кодове ISO 639-1. Въпрос: {question}}} \\
\noindent\textbf{zh:} \langZH{\texttt{\small 预测 3 种对于推理和找到以下事实问题的正确答案最有效的语言。请以 JSON 对象的形式提供回复，其中包含一个键“languages”，其值为 ISO 639-1 语言代码列表。问题：{question}}} \\
\noindent\textbf{hi:} \texttt{\small निम्नलिखित तथ्यात्मक प्रश्न का तर्क देने और सही उत्तर खोजने के लिए सबसे प्रभावी 3 भाषाओं की भविष्यवाणी करें। प्रतिक्रिया को JSON ऑब्जेक्ट के रूप में प्रदान करें जिसमें एकल कुंजी "languages" हो और उसमें ISO 639-1 भाषा कोड की सूची हो। प्रश्न: {question}} \\
\noindent\textbf{bn:} \texttt{\small নিম্নলিখিত তথ্যগত প্রশ্নের যুক্তি দেওয়া এবং সঠিক উত্তর খোঁজার জন্য সবচেয়ে কার্যকরী হবে এমন 3টি ভাষার পূর্বাভাস দিন। প্রতিক্রিয়াটি একটি JSON অবজেক্ট হিসেবে প্রদান করুন যেখানে "languages" নামের একটি মাত্র কী থাকবে এবং তাতে ISO 639-1 ভাষার কোডের তালিকা থাকবে। প্রশ্ন: {question}} \\
\noindent\textbf{ja:} \langJA{\texttt{\small 次の事実に関する質問に対して、推論し正しい答えを見つけるために最も効果的な 3 つの言語を予測してください。応答は、ISO 639-1言語コードのリストを含む単一のキー "languages" を持つJSONオブジェクトとして提供してください。質問：{question}}} \\
\noindent\textbf{id:} \texttt{\small Prediksikan 3 bahasa yang akan paling efektif untuk bernalar dan menemukan jawaban yang benar untuk pertanyaan faktual berikut. Berikan respons sebagai objek JSON dengan satu kunci "languages" yang berisi daftar kode bahasa ISO 639-1. Pertanyaan: {question}} \\
\noindent\textbf{ko:} \langKO{\texttt{\small 다음 사실 관련 질문에 대해 추론하고 정답을 찾는 데 가장 효과적일 3개의 언어를 예측하세요. ISO 639-1 언어 코드 목록을 포함하는 단일 키 "languages"가 있는 JSON 객체로 응답을 제공하세요. 질문: {question}}} \\
\noindent\textbf{pt:} \texttt{\small Preveja 3 idiomas que seriam mais eficazes para raciocinar e encontrar a resposta correta para a seguinte pergunta factual. Forneça a resposta como um objeto JSON com uma única chave "languages" contendo uma lista de códigos de idioma ISO 639-1. Pergunta: {question}} \\
\noindent\textbf{de:} \texttt{\small Sagen Sie 3 Sprachen voraus, die am effektivsten wären, um zu schlussfolgern und die richtige Antwort auf die folgende Sachfrage zu finden. Geben Sie die Antwort als JSON-Objekt mit einem einzigen Schlüssel "languages" an, der eine Liste von ISO 639-1-Sprachcodes enthält. Frage: {question}} \\

\subsection*{Strategy: $L_e \rightarrow L_q$}
\noindent\textbf{en:} \texttt{\small Please think about the following factual question first in English, and then provide your final answer. '{question}'. Provide your final answer in English. The final answer should be concise and directly answer the question. Provide your final answer at the very end after the phrase 'Final Answer:'.} \\
\noindent\textbf{he:} \langHE{\texttt{\small אנא חשוב על השאלה העובדתית הבאה תחילה ב-עברית, ולאחר מכן ספק את תשובתך הסופית. '{question}'. ספק את התשובה הסופית בעברית. התשובה הסופית צריכה להיות תמציתית ולענות ישירות על השאלה. ספק את תשובתך הסופית בסוף לאחר הביטוי 'תשובה סופית:'.}} \\
\noindent\textbf{es:} \texttt{\small Por favor, piensa primero en la siguiente pregunta factual en español y luego proporciona tu respuesta final. '{question}'. Proporciona tu respuesta final en español. La respuesta final debe ser concisa e responder directamente a la pregunta. Proporciona tu respuesta final al final después de la frase 'Respuesta final:'.} \\
\noindent\textbf{fr:} \texttt{\small Veuillez d'abord réfléchir à la question factuelle suivante en français, puis fournissez votre réponse finale. '{question}'. Fournissez votre réponse finale en français. La réponse finale doit être concise et répondre directement à la question. Fournissez votre réponse finale à la fin après la phrase 'Réponse finale:'.} \\
\noindent\textbf{ar:} \langAR{\texttt{\small يرجى التفكير في السؤال الواقعي التالي أولاً باللغة العربية، ثم تقديم إجابتك النهائية. '{question}'. قدم إجابتك النهائية باللغة العربية. يجب أن تكون الإجابة النهائية موجزة وتجيب مباشرة على السؤال. قدم إجابتك النهائية في النهاية بعد العبارة 'الإجابة النهائية:'.}} \\
\noindent\textbf{ru:} \langRU{\texttt{\small Пожалуйста, сначала подумайте над следующим фактическим вопросом на русском, а затем дайте окончательный ответ. '{question}'. Предоставьте свой окончательный ответ на русском языке. Окончательный ответ должен быть кратким и прямо отвечать на вопрос. Предоставьте свой окончательный ответ в самом конце после фразы 'Окончательный ответ:'.}} \\
\noindent\textbf{it:} \texttt{\small Per favore, pensa prima alla seguente domanda fattuale in italiano, quindi fornisci la tua risposta finale. '{question}'. Fornisci la tua risposta finale in italiano. La risposta finale dovrebbe essere concisa e rispondere direttamente alla domanda. Fornisci la tua risposta finale alla fine dopo la frase 'Risposta finale:'.} \\
\noindent\textbf{uk:} \langUK{\texttt{\small Будь ласка, спочатку подумайте над наступним фактичним питанням мовою українською, а потім надайте остаточну відповідь. '{question}'. Надайте свою остаточну відповідь українською. Остаточна відповідь повинна бути стислою та прямо відповідати на питання. Надайте свою остаточну відповідь в самому кінці після фрази 'Остаточна відповідь:'.}} \\
\noindent\textbf{bg:} \langBG{\texttt{\small Моля, първо помислете върху следния фактически въпрос на български и след това предоставете окончателния си отговор. '{question}'. Предоставете окончателния си отговор на български. Окончателният отговор трябва да бъде кратък и да отговаря директно на въпроса. Предоставете окончателния си отговор в самия край след фразата 'Окончателен отговор:'.}} \\
\noindent\textbf{zh:} \langZH{\texttt{\small 请先用 中文 思考以下事实问题，然后提供最终答案。 '{question}'. 请用中文提供最终答案。 最终答案应简洁并直接回答问题。 在最后，请在以下短语后提供你的最终答案： '最终答案：'.}} \\
\noindent\textbf{hi:} \texttt{\small कृपया पहले हिंदी में निम्नलिखित तथ्यात्मक प्रश्न के बारे में सोचें, और फिर अपना अंतिम उत्तर दें। '{question}'. अपना अंतिम उत्तर हिंदी में दें। अंतिम उत्तर संक्षिप्त होना चाहिए और सीधे प्रश्न का उत्तर देना चाहिए। अंत में, इस वाक्यांश के बाद अपना अंतिम उत्तर दें 'अंतिम उत्तर:'.} \\
\noindent\textbf{bn:} \texttt{\small অনুগ্রহ করে প্রথমে বাংলা-এ নিম্নলিখিত তথ্যগত প্রশ্নটি নিয়ে চিন্তা করুন এবং তারপর আপনার চূড়ান্ত উত্তর দিন। '{question}'. আপনার চূড়ান্ত উত্তর বাংলায় দিন। চূড়ান্ত উত্তরটি সংক্ষিপ্ত হওয়া উচিত এবং সরাসরি প্রশ্নের উত্তর দেওয়া উচিত। একেবারে শেষে, এই শব্দগুচ্ছের পরে আপনার চূড়ান্ত উত্তর দিন 'চূড়ান্ত উত্তর:'.} \\
\noindent\textbf{ja:} \langJA{\texttt{\small まず 日本語 で次の事実に関する質問について考え、それから最終的な回答を提供してください。 '{question}'. 最終的な回答を日本語で提供してください。 最終的な回答は簡潔で、質問に直接答える必要があります。 最後に、次のフレーズの後に最終回答を提供してください： '最終回答：'.}} \\
\noindent\textbf{id:} \texttt{\small Tolong pikirkan pertanyaan faktual berikut terlebih dahulu dalam Indonesia, lalu berikan jawaban akhir Anda. '{question}'. Berikan jawaban akhir Anda dalam bahasa Indonesia. Jawaban akhir harus ringkas dan langsung menjawab pertanyaan. Berikan jawaban akhir Anda di bagian paling akhir setelah frasa 'Jawaban akhir:'.} \\
\noindent\textbf{ko:} \langKO{\texttt{\small 먼저 한국어 (으)로 다음 사실 질문에 대해 생각한 다음 최종 답변을 제공하십시오. '{question}'. 최종 답변을 한국어로 제공하세요. 최종 답변은 간결하고 질문에 직접 답해야 합니다. 맨 마지막에 다음 문구 뒤에 최종 답변을 제공하십시오. '최종 답변:'.}} \\
\noindent\textbf{pt:} \texttt{\small Por favor, pense primeiro na seguinte pergunta factual em português e depois forneça a sua resposta final. '{question}'. Forneça sua resposta final em português. A resposta final deve ser concisa e responder directamente à pergunta. Forneça sua resposta final no final após a frase 'Resposta final:'.} \\
\noindent\textbf{de:} \texttt{\small Bitte denken Sie zuerst in Deutsch über die folgende Sachfrage nach und geben Sie dann Ihre endgültige Antwort. '{question}'. Geben Sie Ihre endgültige Antwort auf Deutsch. Die endgültige Antwort sollte prägnant sein und die Frage direkt beantworten. Geben Sie Ihre endgültige Antwort ganz am Ende nach dem Satz an 'Endgültige Antwort:'.} \\

\subsection*{Aggregation - Majority Vote}
\noindent\textbf{en:} \texttt{\small Question: '{question}' Here are answers to the question in different languages: {responses} Based on these responses, determine the correct answer by performing majority voting. Identify the answer that appears most frequently among the provided options and choose it as the correct answer. Provide your reasoning briefly, and then provide your final answer concisely at the very end after the phrase 'Final Answer:'.} \\
\noindent\textbf{he:} \langHE{\texttt{\small שאלה: '{question}' להלן תשובות לשאלה בשפות שונות: {responses} בהתבסס על תשובות אלו, קבע מהי התשובה הנכונה באמצעות הצבעת רוב. זהה את התשובה המופיעה בתדירות הגבוהה ביותר מבין האפשרויות שסופקו ובחר בה כתשובה הנכונה. ספק את הנימוק שלך בקצרה, ולאחר מכן ספק את התשובה הסופית שלך בקצרה בסוף לאחר הביטוי 'תשובה סופית:'.}} \\
\noindent\textbf{es:} \texttt{\small Pregunta: '{question}' Aquí hay respuestas a la pregunta en diferentes idiomas: {responses} Basándote en estas respuestas, determina la respuesta correcta mediante una votación por mayoría. Identifica la respuesta que aparece con más frecuencia entre las opciones proporcionadas y elígela como la respuesta correcta. Proporciona tu razonamiento brevemente y luego proporciona tu respuesta final de forma concisa al final después de la frase 'Respuesta final:'.} \\
\noindent\textbf{fr:} \texttt{\small Question : '{question}' Voici les réponses à la question dans différentes langues : {responses} Sur la base de ces réponses, déterminez la réponse correcte en effectuant un vote à la majorité. Identifiez la réponse qui apparaît le plus fréquemment parmi les options fournies et choisissez-la comme réponse correcte. Fournissez brièvement votre raisonnement, puis fournissez votre réponse finale de manière concise à la fin après la phrase 'Réponse finale:'.} \\
\noindent\textbf{ar:} \langAR{\texttt{\small سؤال: '{question}' فيما يلي إجابات على السؤال بلغات مختلفة: {responses} بناءً على هذه الإجابات، حدد الإجابة الصحيحة من خلال إجراء تصويت بالأغلبية. حدد الإجابة التي تظهر بشكل متكرر بين الخيارات المقدمة واخترها كإجابة صحيحة. قدم تبريرك بإيجاز، ثم قدم إجابتك النهائية باختصار في النهاية بعد العبارة 'الإجابة النهائية:'.}} \\
\noindent\textbf{ru:} \langRU{\texttt{\small Вопрос: '{question}' Вот ответы на вопрос на разных языках: {responses} На основе этих ответов определите правильный ответ методом большинства голосов. Найдите ответ, который встречается чаще всего среди предложенных вариантов, и выберите его в качестве правильного. Кратко приведите свои рассуждения, а затем кратко укажите окончательный ответ в самом конце после фразы 'Окончательный ответ:'.}} \\
\noindent\textbf{it:} \texttt{\small Domanda: '{question}' Ecco le risposte alla domanda in lingue diverse: {responses} In base a queste risposte, determina la risposta corretta eseguendo una votazione a maggioranza. Identifica la risposta che appare più frequentemente tra le opzioni fornite e sceglila come risposta corretta. Fornisci brevemente il tuo ragionamento, quindi fornisci la tua risposta finale in modo conciso alla fine dopo la frase 'Risposta finale:'.} \\
\noindent\textbf{uk:} \langUK{\texttt{\small Питання: '{question}' Ось відповіді на питання різними мовами: {responses} На основі цих відповідей визначте правильну відповідь шляхом голосування більшістю. Визначте відповідь, яка зустрічається найчастіше серед наданих варіантів, і оберіть її як правильну відповідь. Стисло наведіть своє обґрунтування, а потім надайте остаточну відповідь стисло в самому кінці після фрази 'Остаточна відповідь:'.}} \\
\noindent\textbf{bg:} \langBG{\texttt{\small Въпрос: '{question}' Ето отговори на въпроса на различни езици: {responses} Въз основа на тези отговори определете правилния отговор чрез гласуване с мнозинство. Идентифицирайте отговора, който се появява най-често сред предоставените опции, и го изберете за правилен отговор. Представете кратко мотивите си, а след това предоставете окончателния си отговор кратко в самия край след фразата 'Окончателен отговор:'.}} \\
\noindent\textbf{zh:} \langZH{\texttt{\small 问题：'{question}' 以下是该问题的不同语言版本的回答： {responses} 根据这些回答，通过多数投票确定正确答案。找出提供的选项中出现频率最高的答案，并将其选为正确答案。简要说明你的理由，然后在最后，在 '最终答案：' 之后简明扼要地提供你的最终答案。}} \\
\noindent\textbf{hi:} \texttt{\small प्रश्न: '{question}' यहाँ विभिन्न भाषाओं में प्रश्न के उत्तर दिए गए हैं: {responses} इन उत्तरों के आधार पर, बहुमत मतदान करके सही उत्तर निर्धारित करें। उन विकल्पों में से उस उत्तर की पहचान करें जो सबसे अधिक बार दिखाई देता है और उसे सही उत्तर के रूप में चुनें। अपना तर्क संक्षेप में दें, और फिर अंत में, 'अंतिम उत्तर:' वाक्यांश के बाद अपना अंतिम उत्तर संक्षेप में दें।} \\
\noindent\textbf{bn:} \texttt{\small प्रश्न: '{question}' এখানে বিভিন্ন ভাষায় প্রশ্নের উত্তর দেওয়া হলো: {responses} এই উত্তরগুলোর ওপর ভিত্তি করে, সংখ্যাগরিষ্ঠ ভোটের মাধ্যমে সঠিক উত্তরটি নির্ধারণ করুন। প্রদত্ত বিকল্পগুলোর মধ্যে যে উত্তরটি সবচেয়ে বেশিবার এসেছে তা চিহ্নিত করুন এবং সেটিকে সঠিক উত্তর হিসেবে বেছে নিন। সংক্ষেপে আপনার যুক্তি প্রদান করুন এবং সবশেষে, 'চূড়ান্ত উত্তর:' বাক্যাংশের পরে সংক্ষেপে আপনার চূড়ান্ত উত্তর দিন।} \\
\noindent\textbf{ja:} \langJA{\texttt{\small 質問：'{question}' 以下は、さまざまな言語による質問への回答です： {responses} これらの回答に基づいて、多数決により正しい答えを決定してください。提供された選択肢の中で最も頻繁に現れる回答を特定し、それを正解として選択してください。理由を簡単に述べた後、最後に '最終回答：' というフレーズの後に、最終的な答えを簡潔に記述してください。}} \\
\noindent\textbf{id:} \texttt{\small Pertanyaan: '{question}' Berikut adalah jawaban-jawaban untuk pertanyaan tersebut dalam berbagai bahasa: {responses} Berdasarkan jawaban-jawaban ini, tentukan jawaban yang benar dengan melakukan pemungutan suara terbanyak. Identifikasi jawaban yang paling sering muncul di antara pilihan yang tersedia dan pilih sebagai jawaban yang benar. Berikan penalaran Anda secara singkat, lalu berikan jawaban akhir Anda secara ringkas di bagian paling akhir setelah frasa 'Jawaban akhir:'.} \\
\noindent\textbf{ko:} \langKO{\texttt{\small 질문: '{question}' 다음은 여러 언어로 된 질문에 대한 답변입니다: {responses} 이 답변들을 바탕으로 다수결을 통해 정답을 결정하십시오. 제공된 옵션 중에서 가장 자주 나타나는 답변을 식별하고 이를 정답으로 선택하십시오. 추론을 간략하게 제공한 다음, 마지막에 '최종 답변:'라는 문구 뒤에 최종 답변을 간결하게 제공하십시오.}} \\
\noindent\textbf{pt:} \texttt{\small Pergunta: '{question}' Aqui estão as respostas à pergunta em diferentes idiomas: {responses} Com base nessas respostas, determine a resposta correta realizando uma votação por maioria. Identifique a resposta que aparece com mais frequência entre as opções fornecidas e escolha-a como a resposta correta. Forneça seu raciocínio brevemente e, em seguida, forneça sua resposta final de forma concisa bem no final, após a frase 'Resposta final:'.} \\
\noindent\textbf{de:} \texttt{\small Frage: '{question}' Hier sind Antworten auf die Frage in verschiedenen Sprachen: {responses} Bestimmen Sie anhand dieser Antworten die richtige Antwort durch Mehrheitsentscheidung. Identifizieren Sie die Antwort, die unter den angegebenen Optionen am häufigsten vorkommt, und wählen Sie diese als richtige Antwort. Geben Sie Ihre Begründung kurz an und geben Sie dann Ihre endgültige Antwort kurz am Ende nach dem Satz 'Endgültige Antwort:' an.} \\

\subsection*{Aggregation - Minority Aware}
\noindent\textbf{en:} \texttt{\small Question: '{question}' Here are answers to the question in different languages: {responses} Based on these responses, determine the correct answer. Consider all the provided answers and justify your selection. Note that the correct answer is not necessarily the one chosen by the majority of the responses. Keep in mind that knowledge related to this question might be more common or accurate in specific languages. Provide your justification first, and then provide your final answer concisely at the very end after the phrase 'Final Answer:'.} \\
\noindent\textbf{he:} \langHE{\texttt{\small שאלה: '{question}' להלן תשובות לשאלה בשפות שונות: {responses} בהתבסס על תשובות אלו, קבע מהי התשובה הנכונה. שקול את כל התשובות שסופקו ונמק את בחירתך. שים לב שהתשובה הנכונה אינה בהכרח זו שנבחרה על ידי רוב התשובות. זכור שידע הקשור לשאלה זו עשוי להיות נפוץ או מדויק יותר בשפות מסוימות. ספק את הנימוק שלך תחילה, ולאחר מכן ספק את התשובה הסופית שלך בקצרה בסוף לאחר הביטוי 'תשובה סופית:'.}} \\
\noindent\textbf{es:} \texttt{\small Pregunta: '{question}' Aquí hay respuestas a la pregunta en diferentes idiomas: {responses} Basándote en estas respuestas, determina la respuesta correcta. Considera todas las respuestas proporcionadas y justifica tu selección. Ten en cuenta que la respuesta correcta no es necesariamente la elegida por la mayoría de las respuestas. Recuerda que el conocimiento relacionado con esta pregunta puede ser más común o preciso en idiomas específicos. Proporciona tu justificación primero y luego proporciona tu respuesta final de forma concisa al final después de la frase 'Respuesta final:'.} \\
\noindent\textbf{fr:} \texttt{\small Question : '{question}' Voici les réponses à la question dans différentes langues : {responses} Sur la base de ces réponses, déterminez la réponse correcte. Considérez toutes les réponses fournies et justifiez votre sélection. Notez que la réponse correcte n'est pas nécessairement celle choisie par la majorité des réponses. Gardez à l'esprit que les connaissances liées à cette question peuvent être plus courantes ou précises dans des langues spécifiques. Fournissez d'abord votre justification, puis fournissez votre réponse finale de manière concise à la fin après la phrase 'Réponse finale:'.} \\
\noindent\textbf{ar:} \langAR{\texttt{\small سؤال: '{question}' فيما يلي إجابات على السؤال بلغات مختلفة: {responses} بناءً على هذه الإجابات، حدد الإجابة الصحيحة. خذ في الاعتبار جميع الإجابات المقدمة وبرر اختيارك. لاحظ أن الإجابة الصحيحة ليست بالضرورة هي التي اختارتها غالبية الردود. تذكر أن المعرفة المتعلقة بهذا السؤال قد تكون أكثر شيوعاً أو دقة في لغات محددة. قدم تبريرك أولاً، ثم قدم إجابتك النهائية باختصار في النهاية بعد العبارة 'الإجابة النهائية:'.}} \\
\noindent\textbf{ru:} \langRU{\texttt{\small Вопрос: '{question}' Вот ответы на вопрос на разных языках: {responses} Основываясь на этих ответах, определите правильный ответ. Рассмотрите все представленные ответы и обоснуйте свой выбор. Обратите внимание, что правильный ответ не обязательно совпадает с тем, который выбрало большинство. Имейте в виду, что знания по этому вопросу могут быть более полными или точными на определенных языках. Сначала приведите обоснование, а затем кратко укажите окончательный ответ в самом конце после фразы 'Окончательный ответ:'.}} \\
\noindent\textbf{it:} \texttt{\small Domanda: '{question}' Ecco le risposte alla domanda in lingue diverse: {responses} In base a queste risposte, determina la risposta corretta. Considera tutte le risposte fornite e giustifica la tua scelta. Nota che la risposta corretta non è necessariamente quella scelta dalla maggioranza delle risposte. Tieni presente che la conoscenza relativa a questa domanda potrebbe essere più comune o accurata in lingue specifiche. Fornisci prima la tua giustificazione, quindi fornisci la tua risposta finale in modo conciso alla fine dopo la frase 'Risposta finale:'.} \\
\noindent\textbf{uk:} \langUK{\texttt{\small Питання: '{question}' Ось відповіді на питання різними мовами: {responses} На основі цих відповідей визначте правильну відповідь. Розгляньте всі надані відповіді та обґрунтуйте свій вибір. Зауважте, що правильна відповідь не обов'язково є тією, яку обрала більшість. Майте на увазі, що знання, пов'язані з цим питанням, можуть бути більш поширеними або точними в певних мовах. Спочатку надайте своє обґрунтування, а потім стисло надайте остаточну відповідь у самому кінці після фрази 'Остаточна відповідь:'.}} \\
\noindent\textbf{bg:} \langBG{\texttt{\small Въпрос: '{question}' Ето отговори на въпроса на различни езици: {responses} Въз основа на тези отговори определете правилния отговор. Разгледайте всички предоставени отговори и обосновете избора си. Имайте предвид, че правилният отговор не е задължително този, избран от мнозинството. Не забравяйте, че знанията, свързани с този въпрос, могат да бъдат по-често срещани или точни на определени езици. Първо представете обосновката си, а след това предоставете окончателния си отговор кратко в самия край след фразата 'Окончателен отговор:'.}} \\
\noindent\textbf{zh:} \langZH{\texttt{\small 问题：'{question}' 以下是该问题的不同语言版本的回答： {responses} 根据这些回答，确定正确答案。考虑所有提供的答案并说明选择的理由。请注意，正确答案并不一定是大多数回答所选择的那个。请记住，与此问题相关的知识在特定语言中可能更常见或更准确。请先提供你的理由，然后在最后，在 '最终答案：' 之后简明扼要地提供你的最终答案。}} \\
\noindent\textbf{hi:} \texttt{\small प्रश्न: '{question}' यहाँ विभिन्न भाषाओं में प्रश्न के उत्तर दिए गए हैं: {responses} इन उत्तरों के आधार पर, सही उत्तर निर्धारित करें। सभी दिए गए उत्तरों पर विचार करें और अपने चयन का औचित्य साबित करें। ध्यान दें कि सही उत्तर वह नहीं है जो अधिकांश उत्तरों द्वारा चुना गया है। ध्यान रखें कि इस प्रश्न से संबंधित ज्ञान विशिष्ट भाषाओं में अधिक सामान्य या सटीक हो सकता है। पहले अपना औचित्य दें, और फिर अंत में, 'अंतिम उत्तर:' वाक्यांश के बाद अपना अंतिम उत्तर संक्षेप में दें।} \\
\noindent\textbf{bn:} \texttt{\small প্রশ্ন: '{question}' এখানে বিভিন্ন ভাষায় প্রশ্নের উত্তর দেওয়া হলো: {responses} এই উত্তরগুলোর ওপর ভিত্তি করে, সঠিক উত্তরটি নির্ধারণ করুন। প্রদত্ত সমস্ত উত্তর বিবেচনা করুন এবং আপনার নির্বাচনের যৌক্তিকতা ব্যাখ্যা করুন। মনে রাখবেন যে সঠিক উত্তরটি অবশ্যই সংখ্যাগরিষ্ঠ উত্তর দ্বারা নির্বাচিত উত্তর নয়। মনে রাখবেন যে এই প্রশ্ন সম্পর্কিত জ্ঞান নির্দিষ্ট ভাষায় আরও সাধারণ বা সঠিক হতে পারে। প্রথমে আপনার যুক্তি প্রদান করুন এবং সবশেষে, 'চূড়ান্ত উত্তর:' বাক্যাংশের পরে সংক্ষেপে আপনার চূড়ান্ত উত্তর দিন।} \\
\noindent\textbf{ja:} \langJA{\texttt{\small 質問：'{question}' 以下は、さまざまな言語による質問への回答です： {responses} これらの回答に基づいて、正しい答えを決定してください。提供されたすべての回答を検討し、選択の理由を説明してください。正解は必ずしも大多数の回答で選ばれたものとは限らないことに注意してください。この質問に関連する知識は、特定の言語でより一般的または正確である可能性があることを念頭に置いてください。最初に正当化の理由を述べ、最後に '最終回答：' というフレーズの後に、最終的な答えを簡潔に記述してください。}} \\
\noindent\textbf{id:} \texttt{\small Pertanyaan: '{question}' Berikut adalah jawaban-jawaban untuk pertanyaan tersebut dalam berbagai bahasa: {responses} Berdasarkan jawaban-jawaban ini, tentukan jawaban yang benar. Pertimbangkan semua jawaban yang diberikan dan berikan alasan atas pilihan Anda. Perhatikan bahwa jawaban yang benar tidak selalu jawaban yang dipilih oleh mayoritas tanggapan. Ingatlah bahwa pengetahuan yang terkait dengan pertanyaan ini mungkin lebih umum atau akurat dalam bahasa tertentu. Berikan alasan Anda terlebih dahulu, lalu berikan jawaban akhir Anda secara ringkas di bagian paling akhir setelah frasa 'Jawaban akhir:'.} \\
\noindent\textbf{ko:} \langKO{\texttt{\small 질문: '{question}' 다음은 여러 언어로 된 질문에 대한 답변입니다: {responses} 이 답변들을 바탕으로 정답을 결정하십시오. 제공된 모든 답변을 고려하고 선택 이유를 설명하십시오. 정답이 반드시 다수의 답변이 선택한 답변인 것은 아닙니다. 이 질문과 관련된 지식은 특정 언어에서 더 일반적이거나 정확할 수 있음을 명심하십시오. 먼저 정당성을 제공한 다음, 마지막에 '최종 답변:'라는 문구 뒤에 최종 답변을 간결하게 제공하십시오.}} \\
\noindent\textbf{pt:} \texttt{\small Pergunta: '{question}' Aqui estão as respostas à pergunta em diferentes idiomas: {responses} Com base nessas respostas, determine a resposta correta. Considere todas as respostas fornecidas e justifique sua seleção. Note que a resposta correta não é necessariamente a escolhida pela maioria das respostas. Tenha em mente que o conhecimento relacionado a esta pergunta pode ser mais comum ou preciso em idiomas específicos. Forneça sua justificativa primeiro e, em seguida, forneça sua resposta final de forma concisa bem no final, após a frase 'Resposta final:'.} \\
\noindent\textbf{de:} \texttt{\small Frage: '{question}' Hier sind Antworten auf die Frage in verschiedenen Sprachen: {responses} Bestimmen Sie anhand dieser Antworten die richtige Antwort. Berücksichtigen Sie alle bereitgestellten Antworten und begründen Sie Ihre Auswahl. Beachten Sie, dass die richtige Antwort nicht unbedingt diejenige ist, die von der Mehrheit der Antworten gewählt wurde. Denken Sie daran, dass Wissen zu dieser Frage in bestimmten Sprachen verbreiteter oder genauer sein kann. Geben Sie zuerst Ihre Begründung an und geben Sie dann Ihre endgültige Antwort kurz am Ende nach dem Satz 'Endgültige Antwort:' an.} \\

\section{ECLeKTic Pareto Frontier}
\label{sec:appendix_eclektic_budget}

\begin{figure}[ht]
    \centering
    \includegraphics[width=\linewidth]{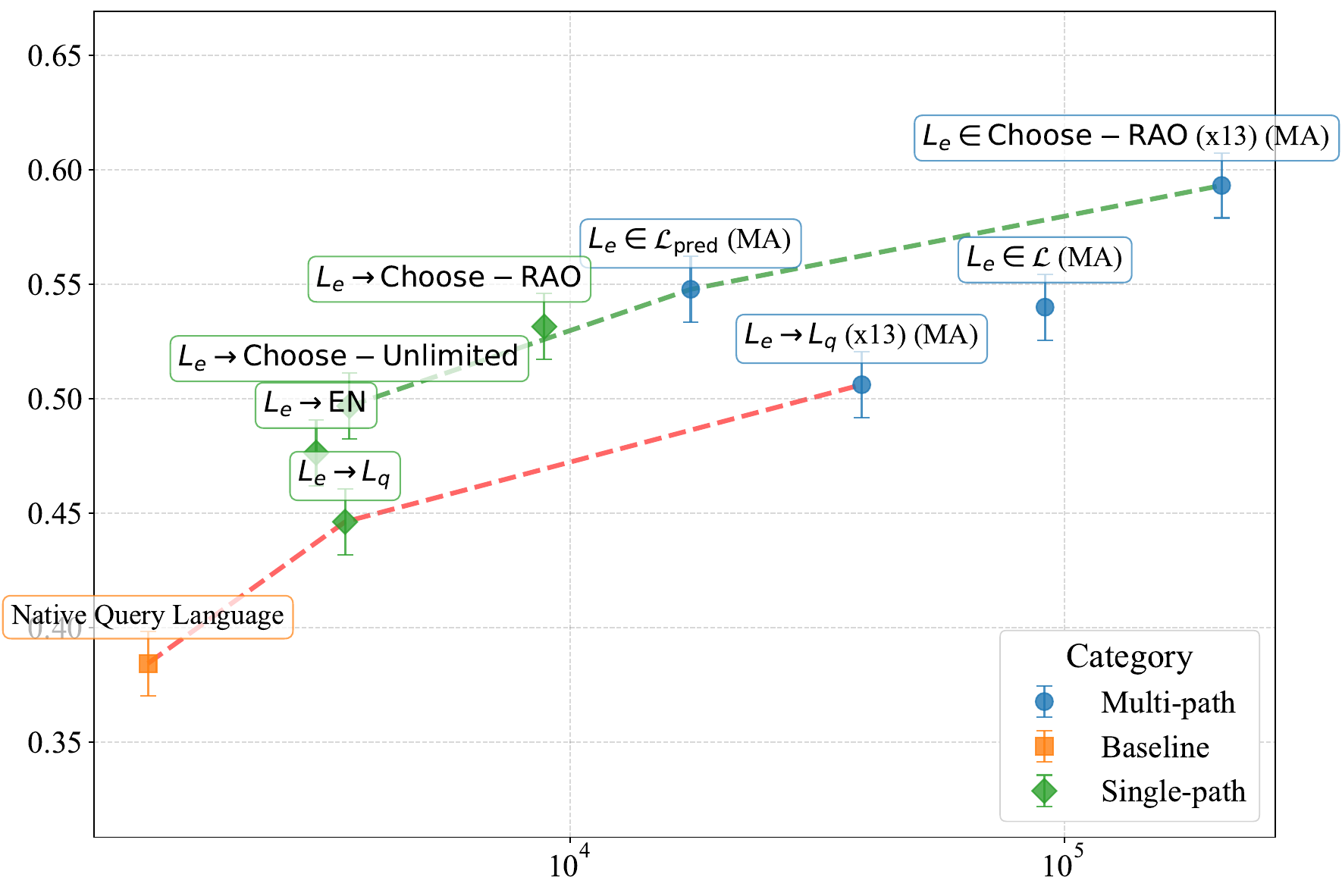}
    \caption{Accuracy versus average token cost (log scale) on the ECLeKTic benchmark. Similar to CLIKE, cross-lingual exploration provides a more efficient compute frontier than native-language scaling.}
    \label{fig:eclektic_budget}
\end{figure}
\section{Artifacts and Data Characteristics}
\label{sec:appendix_artifacts}
The datasets used in this work (ECLeKTic and CLIKE) are derived from public knowledge bases (Wikipedia and Wikidata). Their use in evaluating cross-lingual parametric knowledge is consistent with their intended research purposes. As these datasets primarily concern general factual knowledge and public entities, they are not expected to contain personally identifying information (PII) beyond what is already public, nor do they contain offensive content.

\section{Use of AI Assistants}
\label{sec:appendix_ai_assistants}
During the preparation of this manuscript and the codebase, standard AI assistants (such as GitHub Copilot and Gemini) were used for code generation and formatting in accordance with the ACL policy on AI assistance.

\newpage
\end{document}